\documentclass{article}

\PassOptionsToPackage{numbers, compress}{natbib}
\usepackage[main, final]{neurips_2026}

\usepackage[utf8]{inputenc} 
\usepackage[T1]{fontenc}    
\usepackage{hyperref}       
\usepackage{url}            
\usepackage{booktabs}       
\usepackage{amsfonts}       
\usepackage{nicefrac}       
\usepackage{microtype}      
\usepackage{xcolor}         

\usepackage{multirow}
\usepackage{adjustbox}
\usepackage{wrapfig}
\usepackage{caption}
\usepackage{algorithm}
\usepackage{algorithmic}
\usepackage{amsmath,bm,amsthm,amssymb}
\usepackage{cleveref}

\newcommand{\Crefsub}[2]{%
    \nameCref{#1}~\hyperref[#1]{\ref*{#1}#2}%
}

\hypersetup{
    colorlinks,
    linkcolor={red!50!black},
    citecolor={red!50!black},
    urlcolor={red!50!black}
}

\usepackage{tikz}
\usepackage{pgfplots}
\pgfplotsset{compat=1.18}
\usetikzlibrary{decorations.markings, arrows, arrows.meta, pgfplots.groupplots, positioning, shapes.geometric, matrix, fit, patterns}
\usepgfplotslibrary{statistics}

\definecolor{or}{RGB}{220,140,80}
\definecolor{gr}{RGB}{40,140,80}
\definecolor{bl}{RGB}{70,70,240}
\definecolor{gray}{RGB}{45,45,45}
\definecolor{yl}{RGB}{250,170,30}
\definecolor{sky}{RGB}{100,180,240}
\definecolor{pp}{RGB}{200,150,240}
\definecolor{dr}{RGB}{200,30,0}
\definecolor{dp}{RGB}{225,110,150}
\definecolor{dg}{RGB}{110,175,70}
\definecolor{db}{RGB}{40,40,210}

\tikzstyle{gate} = [red, mark=*, mark size=1.1pt]
\tikzstyle{kvzip} = [gray, mark=x, mark size=2pt, densely dashed, mark options={solid}]
\tikzstyle{duo} = [or, mark=diamond, mark size=1.6pt]
\tikzstyle{snap} = [gr, mark=triangle, mark size=1.6pt]
\tikzstyle{expect} = [bl, mark=+, mark size=1.8pt]

\DeclareMathSymbol{\shortminus}{\mathbin}{AMSa}{"39}
\newcommand{\minus}{\raisebox{.5pt}{$\shortminus$}}

\title{Fast KVzip: Efficient and Accurate LLM Inference with Gated KV Eviction}

\author{%
  Jang-Hyun Kim\thanks{Currently at Apple. E-mail: kimjanghyun1230@gmail.com, sangdoo.yun@navercorp.com} \\
  NAVER AI Lab\\
  \And
  Dongyoon Han \\
  NAVER AI Lab\\
  \And
  Sangdoo Yun \\
  NAVER AI Lab\\
}

\begin{document}

\maketitle

\begin{abstract}
Efficient key-value (KV) cache management is crucial for the practical deployment of large language models (LLMs), yet existing compression techniques often incur a trade-off between performance degradation and computational overhead. We propose a novel gating-based KV cache eviction method for frozen-weight LLMs that achieves high compression ratios with negligible computational cost. Our approach introduces lightweight sink-attention gating modules to identify and retain critical KV pairs, and integrates seamlessly into both the prefill and decoding stages. The proposed gate training algorithm relies on forward passes of an LLM, avoiding expensive backpropagation, while achieving strong task generalization through a task-agnostic reconstruction objective. Extensive experiments across the Qwen2.5-1M, Qwen3, and Gemma3 families show that our method maintains near-lossless performance while evicting up to 70\% of the KV cache. The results are consistent across a wide range of tasks, including long-context understanding, code comprehension, and mathematical reasoning, demonstrating the generality of our approach. 
Project page: \href{https://janghyun1230.github.io/fastkvzip/}{https://janghyun1230.github.io/fastkvzip/}
\end{abstract}

\section{Introduction}

Transformer-based LLMs have revolutionized the landscape of artificial intelligence, solving complex challenges ranging from long-context processing to advanced reasoning \citep{o1,gemini}.
A cornerstone of this success is the attention mechanism, which allows models to effectively contextualize features across sequences \citep{transformer}.
To mitigate the computational cost of re-calculating attention features at every step, inference systems adopt KV caching \citep{transformerxl}.

\begin{wrapfigure}[15]{r}{0.5\textwidth}
    \centering
    \vspace{-1.4em}
    \resizebox{1.04\linewidth}{!}{%
    \begin{tikzpicture}

    \begin{groupplot}[group style={columns=2, horizontal sep=1.2cm, vertical sep=0.0cm}]
        \nextgroupplot[
            width=4.3cm,
            height=3.8cm,
            every axis plot/.append style={thick},
            grid=major,
            xmajorgrids=true,
            ymajorgrids=true,
            major grid style={dotted, black},
            xlabel={KV cache budget ratio},
            ylabel={Accuracy (\%)},
            xlabel shift=-0.15cm,
            ylabel shift=-0.15cm,
            xlabel near ticks,
            ylabel near ticks,
            label style={font=\footnotesize},
            tick label style={font=\scriptsize},
            tick pos=left,
            xtick={0.2, 0.5, 0.75, 1.0},
            extra x ticks={0.3, 0.4},   
            extra x tick labels={,},
            extra x tick style={
                grid=none,
                tick style={thin},
            },    
            y tick label style={/pgf/number format/.cd, fixed, fixed zerofill, precision=0},
            xmin=0.2,
            xmax=1.0,
            ymax=30.0,
            ymin=10.0,
            ytick distance=5,
            legend image post style={scale=0.6},
            legend style={legend columns=1, font=\scriptsize, at={(0.98,0.04)}, anchor=south east, inner sep=1pt, row sep=-2pt},
            legend cell align={left},
            title={(a) Performance},
            title style={font=\small, xshift=-0.5em, yshift=-0.3em},
        ]

        \addplot[gate] table[x=ratio, y=gate-mrcr, col sep=comma]{data/qwen2.5-7b.csv};\addlegendentry{Fast KVzip}
        \addplot[snap] table[x=ratio, y=kvzip-mrcr, col sep=comma]{data/qwen2.5-7b.csv};\addlegendentry{KVzip}
        \addplot[expect] table[x=ratio, y=snap-mrcr, col sep=comma]{data/qwen2.5-7b.csv};\addlegendentry{SnapKV}

        \nextgroupplot[
            width=4.3cm,
            height=3.8cm,
            every axis plot/.append style={thick},
            grid=major,
            xmajorgrids=true,
            ymajorgrids=true,
            major grid style={dotted, black},
            xlabel={Peak KV Memory (GB)},
            ylabel={Prefill Time (s)},
            xlabel shift=-0.15cm,
            ylabel shift=-0.15cm,
            xlabel near ticks,
            ylabel near ticks,
            label style={font=\footnotesize},
            tick label style={font=\scriptsize},
            tick pos=left,
            ymin=0,
            ytick distance=30,
            xmin=0,
            xmax=36,
            xtick distance=6,
            title={(b) Efficiency},
            title style={font=\small, xshift=-0.5em, yshift=-0.3em},
        ]

        \addplot+[only marks, mark=*, mark size=1.6pt, mark options={fill=gr,draw=gr}] coordinates {(33.2, 152)};      
        \addplot+[only marks, mark=*, mark size=1.6pt, mark options={fill=black!80,draw=black!80}] coordinates {(33.2, 48)};     
        \addplot+[only marks, mark=*, mark size=1.6pt, mark options={fill=red,draw=red}] coordinates {(12.2, 25)};   

        \node[anchor=south east, font=\scriptsize] at (axis cs:34,120) {KVzip};
        \node[anchor=south east, font=\scriptsize] at (axis cs:35,48) {No Compression};
        \node[anchor=south east, font=\scriptsize] at (axis cs:18,25) {Fast KVzip};

    \end{groupplot}
\end{tikzpicture}
    }
    \vspace{-1em}
    \caption{\textbf{Highlighted results.} (a) KV cache compression performance on OpenAI MRCR \citep{mrcr} compared to state-of-the-art methods using Qwen2.5-7b-1M \citep{qwen}. (b) Chunked-prefill efficiency for 170K tokens on Qwen2.5-7b-14M (BF16 precision) with 30\% KV cache budget.}
    \label{graph:intro}
\end{wrapfigure}

However, the effectiveness of KV caching is increasingly strained by the growing demand for long-context inference. As the memory required to store KV features scales linearly with sequence length, KV caches become a dominant bottleneck during inference \citep{vllm}.
While KV cache eviction approaches alleviate this burden, they often face a critical dilemma: Methods with negligible compression overhead result in significant performance degradation \citep{h2o,snapkv}, while methods that preserve accuracy tend to incur prohibitive compression overhead \citep{kvzip}.
For instance, KVzip \citep{kvzip} relies on the reconstruction process that doubles computation during prefill (see \Crefsub{graph:intro}{b}), limiting its practicality for latency-sensitive deployment.

In this paper, we introduce \textbf{\textit{Fast KVzip}}, a fast and accurate KV cache eviction method that addresses the inefficiencies of prior reconstruction-based approaches.
Our key insight is that the future utilization of KV pairs is largely an \textit{intrinsic property} that can be directly decoded from the input hidden states, without reconstructing the entire context as in KVzip. Leveraging this insight, we seamlessly integrate a lightweight gating mechanism into the Transformer’s forward pass. This allows the model to directly assess the importance of KV pairs during both the prefill and decoding stages, incurring negligible memory and computational overhead while preserving high compression fidelity.

To identify an effective gating mechanism, we systematically explore the design space. Our empirical analysis reveals that gates perform most effectively when operating on input hidden states rather than intermediate attention features. Furthermore, we find that an architecture inspired by attention sinks \citep{streaming} yields the strongest performance.
Accordingly, we introduce a layer-wise gate training procedure that distills KV importance scores pre-computed from a compute-intensive reconstruction process. To ensure efficiency and avoid degrading the original model’s capabilities, we keep the LLM weights frozen and train only the gating parameters, requiring less than one H100 GPU hour for 14B-scale models.

As highlighted in \Cref{graph:intro}, Fast KVzip matches the compression performance of KVzip while maintaining model accuracy at KV budget ratios as low as 40\% on OpenAI MRCR \citep{mrcr}. Moreover, \Crefsub{graph:intro}{b} shows that Fast KVzip achieves approximately 5$\times$ speedup in 170K-token prefill over KVzip, while reducing prefill peak memory by roughly 3$\times$ compared to the no-compression baseline.

We further demonstrate the effectiveness and generality of our method across a diverse set of state-of-the-art open-source LLMs, including Qwen2.5-7B/14B-1M \citep{qwen}, Qwen3-8B/14B, Qwen3-8B-FP8 \citep{qwen3}, and Gemma3-12B \citep{gemma3}. Our approach is compatible with quantized weights and sliding-window hybrid attention mechanisms.
We evaluate on prefill-intensive long-context benchmarks such as SCBench \citep{scbench} and MRCR \citep{mrcr} with context lengths of up to 170K tokens, as well as decoding-intensive reasoning tasks including AIME24 \citep{aime} and MATH \citep{math}. Across all models and benchmarks, Fast KVzip achieves near-lossless performance while reducing the KV cache to 30$\minus$40\% of its original size, underscoring the robustness and practicality of the proposed gating-based eviction strategy.
\section{Related Work}

\begin{wrapfigure}[17]{r}{0.5\textwidth}
    \centering
    \vspace{-1.2em}
    \captionof{table}{\textbf{Gating strategies for Transformer attention.} The table below indicates operation execution per layer for a given input. Mixture-of-Depths (MoD) uses gating to skip attention computation, while Mixture-of-Block-Attention (MoBA) applies gating to select blocks for sparse attention with stored KV features. Our method bridges these approaches by conducting an attention operation for every input token while improving computational and memory efficiency through gated KV eviction.}
    \small
    \begin{tabular}{lll}
    \toprule
    Method & Attention & Store KV feature \\
    \midrule
    MoD \citep{mixture-depth} & Conditional  & Conditional \\
    MoBA \citep{moba} & Always & Always \\
    \midrule
    Fast KVzip & Always & Conditional \\
    \bottomrule
    \end{tabular}    
    \label{tab:gating_mechanism}
\end{wrapfigure}

\paragraph{Gating in Transformers.}
To address the scaling issues of Transformers \citep{transformer}, researchers have introduced gating mechanisms that enable sparse and selective computation, including MoE \citep{moe}, MoD \citep{mixture-depth}, and MoBA \citep{moba}. These approaches demonstrate that pretraining with gating can effectively exploit sparsity in MLP and attention modules, improving computational efficiency. \Cref{tab:gating_mechanism} provides a conceptual comparison of these approaches with our method.

In the landscape of KV cache compression, Locret \citep{locret} employs a gating module to predict KV importance scores, while DMS \citep{dms} trains an eviction policy via logit distillation. TrimKV \citep{trimkv} introduces a retention-gated mechanism to learn time-decaying importance scores.
Despite their promise, Locret is confined to prefill-stage compression, and TrimKV suffers from performance degradation in prefill-intensive scenarios, specializing in domains of mathematical reasoning. In our work, we provide a comprehensive interpretation of the gating mechanism, demonstrating that well-designed gating architectures and optimization strategies achieve near-lossless compression across both prefill and decoding stages. We provide a more detailed comparison to prior methods in \Cref{appendix:experiment}.
We also note a concurrent work, KVzap \citep{kvzap}, which explores a similar gating optimization objective but differs in specific architecture and training recipe. Together with our work, these parallel efforts underscore the promise of gating-based approaches for KV cache compression.

\paragraph{KV Cache Compression.}
To mitigate the scalability bottleneck of KV caching in autoregressive Transformers \citep{gpt2}, efficient attention mechanisms have been proposed, including sliding-window attention \citep{sparsetransformer}, linear attention \citep{linear}, latent attention \citep{mla}, and grouped-query attention \citep{gqa}. While effective, these methods typically require training from scratch or substantial fine-tuning to accommodate architectural changes.

Recent work observes inherent sparsity in LLM attention patterns \citep{dejavu}. Methods such as $\text{H}_2\text{O}$ \citep{h2o}, Finch \citep{corallo2024finch}, and SnapKV \citep{snapkv} exploit sparsity observed during prefill or decoding to evict under-utilized KV pairs. However, sparsity patterns inferred at inference time often overfit the current input and fail to generalize across future queries. To address this limitation, query-agnostic approaches, \textit{e.g.,} KVzip \citep{kvzip}, EpiCache \citep{epicache}, and Expected Attention \citep{kvpress}, estimate KV importance independently of specific queries and demonstrate improved robustness in multi-query settings. Orthogonal strategies, such as DuoAttention \citep{duo}, RLKV \citep{rlkv}, and FastGen \citep{fastgen}, predefine attention patterns (e.g., sliding-window or A-shaped) using calibration datasets prior to deployment. Beyond token-level eviction, complementary compression techniques explore alternative dimensions, including feature-channel compression \citep{think}, low-rank compression \citep{palu}, and KV quantization \citep{kivi}.

\section{Method}\label{sec:method}

\begin{figure*}[t]
    \begin{center}


\resizebox{0.98\linewidth}{!}{
\begin{tikzpicture}

\definecolor{darkred}{RGB}{170, 0, 0}
\definecolor{myblue}{RGB}{80, 150, 220}
\definecolor{mygreen}{RGB}{120, 200, 60}

\definecolor{myred}{RGB}{230, 140, 100}
\definecolor{myorange}{RGB}{240, 220, 120}

\tikzstyle{kvcache} = [draw=black, line width=0.4pt, rounded corners=.06cm, minimum width=0.8cm, minimum height=0.18cm, inner sep=0pt]
\tikzstyle{arrow} = [-stealth, line width=0.2 mm]
\tikzstyle{label} = [font=\scriptsize]
\tikzstyle{patterns} = [pattern=north east lines, pattern color=black,draw=none, minimum width=0.2cm, minimum height=0.2cm, inner sep=0pt, anchor=east]

\begin{scope}[shift={(0, 0)}]
    \node[font=\fontsize{7}{9.5}\selectfont, anchor=west] at (-1.15, 2.35) {(a) Gating in Attention Modules};

    \node[label] (hidden) at (0.0, -0.2) {Hidden states};

    \node[kvcache, fill=myorange!40, above=0.8cm of hidden] (kv_pairs) {};

    \node[fill=gray!6, rounded corners=.15cm, minimum width=2.4cm, minimum height=0.6cm, inner sep=0.6cm, right=0.1cm of kv_pairs, yshift=-0.1cm, opacity=0.8] (shared_bg) {};      
    \node[kvcache, fill=myorange!40, right=0.9cm of kv_pairs] (imp_scores) {};
    \node[label, above=-0.04cm of kv_pairs] (qkv) {QKV};
    \node[label, above=-0.04cm of imp_scores] {\textcolor{darkred}{KV importance scores}};

    \draw[arrow] (hidden) -- node[label, left, yshift=0.12cm] {Projection} (kv_pairs);
    \draw[arrow] ($(hidden.north)!0.4!(kv_pairs.south)$) -| node[label, right, yshift=0.2cm] {Gating} (imp_scores);

    \node[label, above=0.6cm of kv_pairs] (output) {Attention outputs};
    \draw[arrow] (qkv) -- ($(output.south) + (0.0,0.1)$);

\end{scope}

\draw[dashed] (3.3, -0.3) -- (3.3, 2.4);

\begin{scope}[shift={(8, 0)}]    
    \node[font=\fontsize{7}{9.5}\selectfont, anchor=west] (title_a) at (-2.8, 2.35) {(b) KV Caching with Eviction};

    \node[kvcache, fill=myblue] (chunki) at (0,0) {};
    \node[label, below=-0.04cm of chunki] {$i$-th chunk};
    
    \node[kvcache, fill=myblue!50, left=1.5cm of chunki] (input_tok) {};
    \node[label, left=0.1cm of input_tok] {Input tokens};
    \node at ($(input_tok)!0.5!(chunki)$) {...};
    \node[right=0.5cm of chunki] {...};
    
    \node[kvcache, fill=myorange!70, above=0.4cm of chunki] (kvfeat) {};
    \node[label, right=0.cm of kvfeat, align=left] {KV w/ \textcolor{darkred}{scores}};
    
    \draw[arrow] (chunki) -- node[label, font=\tiny, midway, right, yshift=-0.03cm] {Model forward} (kvfeat);
    
    \node[rounded corners=.0cm, minimum width=2.26cm, minimum height=0.34cm, fill=gray!10, above=0.17cm of kvfeat, xshift=-0.03cm, draw=gray!50] {};
    
    \node[kvcache, fill=myorange!70, above=0.25cm of kvfeat, xshift=0.9cm] (concat_2) {};
    \node[kvcache, minimum width=1.6cm, fill=myred!60, left=0.0cm of concat_2] (concat_1) {};
    \draw[arrow] (kvfeat.north) to [out=60, in=240, looseness=0.8] (concat_2.south);
    \node[patterns] (patternnode) at (concat_2.east) {};
    
    \node[label, font=\tiny, above=0.05cm of concat_2.east, xshift=0.4cm] {Local window};

    \node[kvcache, minimum width=1.6cm, fill=myred, above=0.8cm of kvfeat] (cache_i_main) {};
    \node[label, above=-0.04cm of cache_i_main] {KV cache ($i$)};
    \node[patterns] (patternnode) at (cache_i_main.east) {};

    \draw[arrow] ($(kvfeat |- concat_2.north) + (0, 0.6mm)$) -- node[label, font=\tiny, midway, right, yshift=0.02cm] {Evict} (cache_i_main);

    \node[kvcache, minimum width=1.6cm, fill=myred!60, left=1.5cm of cache_i_main] (cache_prev) {};
    \node[label, above=-0.04cm of cache_prev] {KV cache ($i-1$)};
    \draw[arrow] (cache_prev.east) .. controls +(0.8,-0.0) and +(-0.4,0.0) .. (concat_1.west);
    \draw[arrow] (cache_prev.east) .. controls +(0.8,-0.0) and +(-2.2,0.0) .. ($(chunki.north)!0.5!(kvfeat.south)$);
    
\end{scope}

\end{tikzpicture}
}

        \vspace{-0.5em}
        \caption{\textbf{Computational flow.} (a) Illustration of gating at each attention layer. (b) During each forward pass, we jointly calculate importance scores and evict low-importance KV pairs. For decoding, we maintain a small buffer to cache recent hidden states, performing gating and eviction in parallel once the buffer is full. This parallelized computation reduces latency overhead. We utilize a chunk size of 16K for prefill and a buffer size of 128 tokens for decoding. Empirically, we observe that retaining a local window of recent KV features at each compression step slightly improves performance (see \Cref{appendix:exp_local_window}), though our method remains robust even without the local window.}
        \label{fig:method}
    \end{center}
\end{figure*}

\subsection{Gating Mechanism}

\paragraph{Formulation.} We design a gating module that processes the hidden features of each attention layer and outputs importance scores for the KV features in that layer. Specifically, the gate at layer $l$ is defined as
$g_l: \mathbb{R}^D \rightarrow [0,1]^H$, 
where $D$ denotes the hidden feature dimension and $H$ is the number of KV heads in the layer. We introduce an additional gating branch into the model’s forward computation. This branch operates independently and does not affect the original attention outputs (see \Crefsub{fig:method}{a}).

\paragraph{Computational Flow.}

\Crefsub{fig:method}{b} illustrates the overall computational flow of our method. We adopt a chunked prefill to reduce the peak memory usage when processing long contexts \citep{prefill}. For each input chunk, we first compute the KV features and their corresponding importance scores using the gating mechanism. Based on these scores, we evict low-importance KV features from the cache, maintaining a compressed KV cache either at a fixed retention ratio or within a predefined memory budget. Our approach is compatible with both non-uniform and uniform KV cache structures \citep{adakv,snapkv}. By maintaining a compressed KV cache throughout chunked prefill, our method reduces the peak KV cache memory compared to naive chunked prefill, as shown in \Crefsub{graph:intro}{b}.

During decoding, we apply the same gated KV eviction strategy. However, computing gates at every decoding step introduces substantial overhead due to frequent function calls and increased I/O, resulting in a 30$\minus$60\% latency increase depending on the context length. To mitigate this issue, we maintain a small buffer of token length 128 that caches recent hidden states and periodically computes gating in parallel. After updating the importance scores, we perform KV compression accordingly. This strategy reduces the gating overhead to approximately 1\% of the model’s forward-pass latency.

\subsection{Interpretation of Gating Formulation}

\begin{wrapfigure}[13]{r}{0.5\textwidth}
    \vspace{-2em}
    \begin{center}


\resizebox{\linewidth}{!}{
\begin{tikzpicture}

\definecolor{darkred}{RGB}{170, 0, 0}
\definecolor{myblue}{RGB}{80, 150, 220}
\definecolor{mygreen}{RGB}{120, 200, 60}

\definecolor{myred}{RGB}{230, 140, 100}
\definecolor{myorange}{RGB}{240, 220, 120}

\tikzstyle{kvcache} = [draw=black, line width=0.4pt, rounded corners=.06cm, minimum width=0.8cm, minimum height=0.2cm, inner sep=0pt]
\tikzstyle{arrow} = [-stealth, line width=0.2 mm]
\tikzstyle{label} = [font=\footnotesize]

\begin{scope}[shift={(0, 0)}]
    \node[label] at (0.7, 2.) {(a) DuoAttention};
    \node[label, anchor=west] (input) at (0., 0.) {Hidden states};
    \node[label, above=0.95cm of input] (output) {$\{0,1\}^H$};
    \draw[arrow] (input) -- node[label, left, align=left] {Constant\\ function} node[label, right] {$g_l$} (output);
\end{scope}

\begin{scope}[shift={(3, 0)}]
    \node[label] at (0.5, 2.) {(b) Expected Attention};
    \node[label, anchor=west] (input) at (0., 0.) {Key states};
    \node[label, above=0.95cm of input] (output) {$[0,1]^H$};
    \draw[arrow] (input) -- node[label, left, align=left] {Quadratic\\ function} node[label, right] {$g_l$} (output);

\end{scope}

\begin{scope}[shift={(5.55, 0)}]
    \node[label] at (0.65, 2.) {(c) FastGen};
    \node[label, anchor=west] (input) at (0., 0.) {Input tokens};
    \node[label, above=0.95cm of input] (output) {$\{0,1\}^H$};
    \draw[arrow] (input) -- node[label, left, align=left] {Boolean\\ function} node[label, right] {$g_l$} (output);
\end{scope}

\end{tikzpicture}
}

        \caption{\textbf{Gating formulation} of existing KV compression methods at each attention layer. Note that we compare the core components of baseline methods, as each method jointly uses other techniques. For instance, FastGen combines its approach with cumulative attention scores \citep{h2o}.
        }
        \label{fig:baseline}
    \end{center}
\end{wrapfigure}

Before detailing our method, we establish a conceptual basis for why gating mechanisms are effective for KV cache compression. Notably, we observe that several prominent compression methods can be reformulated as constrained instances of a generalized gating framework (\Cref{fig:baseline}).

For instance, DuoAttention \citep{duo} optimizes a head-level indicator function to designate specific heads for local-window attention. This represents a constrained gating formulation where the gating function is a constant Boolean mapping, independent of inputs. Expected Attention \citep{kvpress} approximates future attention scores by applying a quadratic model of key states with mean query-state statistics. Furthermore, FastGen \citep{fastgen} identifies heads that attend to specific token clusters (\textit{e.g.}, punctuation tokens), which can be interpreted as a head-wise gating function on input tokens.

Our gating approach generalizes these prior methods by allowing any differentiable gating function without imposing empirical and heuristic constraints on the optimization. This interpretation enables a principled approach to KV cache compression, transforming the problem from heuristic design to data-driven optimization, effectively utilizing the model to discover the optimal compression strategy.


\subsection{Gate Training}\label{subsec:method_optimization}

\begin{wrapfigure}[14]{r}{0.5\textwidth}
    \vspace{-3em}
    \begin{center}
        \input{neurips/figure/optimization}
        \caption{\textbf{Gate training process.} We compute the importance scores of KV pairs for a given text input following KVzip, which calculates the maximum attention score each KV pair receives during the context reconstruction \citep{kvzip}. We train gating modules to predict these scores from the corresponding input hidden states at each attention layer given the text. Please refer to \Cref{appendix:details} for further details and the training hyperparameters.}
        \label{fig:optimization}
    \end{center}
\end{wrapfigure}

\paragraph{Consideration.}
We impose three requirements on our optimization: (1) computational efficiency, (2) robustness to task-specific overfitting, and (3) preservation of the base LLM’s performance. End-to-end optimization via backpropagation, for example, incurs substantial memory overhead, limiting the usable context length and increasing training time. Furthermore, training gating mechanisms on data from a specific domain or task, such as mathematical reasoning \citep{trimkv} or instruction-based question answering \citep{locret}, may yield domain-specific gains but incur degraded performance on other domains or tasks.

\paragraph{Training Procedure.}
With these considerations, we propose a gate training procedure illustrated in \Cref{fig:optimization}. We optimize the gates at each layer using target scores derived from KVzip’s reconstruction process \citep{kvzip}, while freezing LLM weights to avoid altering the model’s inherent capabilities. This design is computationally efficient, as reconstruction can be performed using parallelized forward passes. Given the target scores, we optimize gates independently at each layer using stochastic gradient descent with a binary cross-entropy loss \citep{bce}. The resulting optimization is embarrassingly parallel across both samples and layers, enabling efficient GPU execution.

Reconstruction-based target signals have been shown to generalize well to downstream tasks \citep{bert,kvzip}, thereby mitigating task-overfitting issues. While KVzip introduces additional compression latency at inference time due to reconstruction, we perform this step only during a pre-deployment phase to train the gates. Consequently, inference remains fast. 

\paragraph{Training Data.} 
We train the gating mechanism using the FineWeb-Edu pretraining corpus \citep{fineweb}, which has no overlap with downstream evaluation datasets. We randomly sample sequences ranging from 10K to 30K tokens, yielding a total of 1M training tokens. Given that FineWeb-Edu contains over 1T tokens, our training set corresponds to approximately a $10^{-6}$ fraction of the full corpus. In \Cref{appendix:sensitivity}, we analyze the effect of data source and size to our method, demonstrating its robustness. 

\begin{figure}[t]
    \centering
    \resizebox{1.02\linewidth}{!}{%
\begin{tikzpicture}

    \begin{groupplot}[
        group style={group name=groupa, columns=2, horizontal sep=1.2cm, vertical sep=0.0cm},
        width=4cm, height=3.5cm,
        every axis plot/.append style={thick},
        grid=major, xmajorgrids=true, ymajorgrids=true,
        major grid style={dotted, black},
        xlabel={KV cache ratio}, ylabel={Accuracy (\%)},
        xlabel shift=-0.15cm, ylabel shift=-0.15cm,
        xlabel near ticks, ylabel near ticks,
        label style={font=\footnotesize},
        tick label style={font=\scriptsize},
        tick pos=left,
        xmax=1.0, xmin=0.2,
        xtick={0.2, 0.5, 0.75, 1.0},
        extra x ticks={0.3, 0.4},
        extra x tick labels={,},
        extra x tick style={grid=none, tick style={thin}},
    ]
        \nextgroupplot[
            y tick label style={/pgf/number format/.cd, fixed, fixed zerofill, precision=0},
            ymax=80.0,
            title={Retrieval},
            title style={font=\small, xshift=0em, yshift=-0.3em},
            legend columns=3,
            legend style={at={(1.2, 1.4)}, anchor=south, font=\footnotesize},
        ]
        \addplot[gate] table [y=recon-retv, col sep=comma]{data/opt-target.csv};\addlegendentry{Reconstruction}
        \addplot[snap] table [y=next-retv, col sep=comma]{data/opt-target.csv};\addlegendentry{Next Token}
        \addplot[expect] table [y=qa-retv, col sep=comma]{data/opt-target.csv};\addlegendentry{QA}

        \nextgroupplot[
            y tick label style={/pgf/number format/.cd, fixed, fixed zerofill, precision=0},
            ymax=100.0,
            title={Contextual QA},
            title style={font=\small, xshift=0em, yshift=-0.3em},
        ]
        \addplot[gate] table [y=recon-qa, col sep=comma]{data/opt-target.csv};
        \addplot[snap] table [y=next-qa, col sep=comma]{data/opt-target.csv};
        \addplot[expect] table [y=qa-qa, col sep=comma]{data/opt-target.csv};
    \end{groupplot}
    \node[below, font=\normalsize] at ($(groupa c1r1.south)!0.5!(groupa c2r1.south)+(0,-0.7cm)$) {(a) Effect of target scores};

    \begin{scope}[xshift=7.7cm]
    \begin{groupplot}[
        group style={group name=groupb, columns=2, horizontal sep=1.2cm, vertical sep=0.0cm},
        width=4cm, height=3.5cm,
        every axis plot/.append style={thick},
        grid=major, xmajorgrids=true, ymajorgrids=true,
        major grid style={dotted, black},
        xlabel={KV cache ratio}, ylabel={Accuracy (\%)},
        xlabel shift=-0.15cm, ylabel shift=-0.15cm,
        xlabel near ticks, ylabel near ticks,
        label style={font=\footnotesize},
        tick label style={font=\scriptsize},
        tick pos=left,
        xmax=1.0, xmin=0.2,
        xtick={0.2, 0.5, 0.75, 1.0},
        extra x ticks={0.3, 0.4},
        extra x tick labels={,},
        extra x tick style={grid=none, tick style={thin}},
    ]
        \nextgroupplot[
            y tick label style={/pgf/number format/.cd, fixed, fixed zerofill, precision=0},
            ytick distance=10,
            title={Retrieval},
            title style={font=\small, xshift=0em, yshift=-0.3em},
            legend columns=3,
            legend style={at={(1.1, 1.4)}, anchor=south, font=\footnotesize,
                /pgfplots/legend image code/.code={
                    \draw[mark repeat=2,mark phase=2]
                    plot coordinates {(0cm,0cm) (0.2cm,0cm) (0.4cm,0cm)};
                }},
        ]
        \addplot[gate] table [y=hidden-retv, col sep=comma]{data/opt-input.csv};\addlegendentry{Hidden State}
        \addplot[snap] table [y=key-retv, col sep=comma]{data/opt-input.csv};\addlegendentry{Key State}
        \addplot[expect] table [y=key-prerope-retv, col sep=comma]{data/opt-input.csv};\addlegendentry{Pre-RoPE Key State}

        \nextgroupplot[
            y tick label style={/pgf/number format/.cd, fixed, fixed zerofill, precision=0},
            ymin=90,
            ytick distance=1.0,
            title={Contextual QA},
            title style={font=\small, xshift=0em, yshift=-0.3em},
        ]
        \addplot[gate] table [y=hidden-qa, col sep=comma]{data/opt-input.csv};
        \addplot[snap] table [y=key-qa, col sep=comma]{data/opt-input.csv};
        \addplot[expect] table [y=key-prerope-qa, col sep=comma]{data/opt-input.csv};
    \end{groupplot}
    \node[below, font=\normalsize] at ($(groupb c1r1.south)!0.5!(groupb c2r1.south)+(0,-0.7cm)$) {(b) Effect of gate inputs};
    \end{scope}

\end{tikzpicture}
    }
    \vspace{-1em}
    \caption{\textbf{Gate Training Design Analysis}. 
    We evaluate gate training targets and inputs on Qwen2.5-7B-1M using a retrieval task (SCBench.KV) \citep{scbench} and a contextual QA task (SQuAD) \citep{squad}. In (a), we compare different tasks to obtain target scores for gate training. For a fair comparison, we fix the gate architecture and only vary the task used to compute the maximum attention scores in our gate training (\Cref{fig:optimization}). We use FineWeb-Edu for training data, except for QA where we use LongAlpaca \citep{longalpaca} following \citet{locret}.
    }
    \label{fig:opt-target}
\end{figure}

\begin{wrapfigure}[16]{r}{0.5\textwidth}
    \vspace{-1.5em}
    \centering
    \input{neurips/figure/heatmap}
    \vspace{-1em}
    \caption{\textbf{Maximum attention scores} received by KV features across specific tasks. 
    We visualize layers 7, 14, and 21 of Qwen2.5-7B-1M, with other layers exhibiting similar patterns, using the gate training data from \Crefsub{fig:opt-target}{a}.}
    \label{fig:heatmap}
\end{wrapfigure}

\paragraph{Alternatives.}
In \Crefsub{fig:opt-target}{a}, we compare the reconstruction-based target KV scores with scores derived from next-token prediction \citep{trimkv} and instruction-based question answering \citep{locret}. The figure demonstrates that reconstruction-based targets generalize significantly better, while alternatives lead to inferior performance. 

\Cref{fig:heatmap} reveals that reconstruction-based attention patterns are uniform and structured across KV sequences. In contrast, next-token prediction generates denser patterns, reflecting more intensive feature contextualization. We hypothesize that the reconstruction task is effective because many downstream applications rely primarily on these high-level contextualized features \citep{kvzip}. To maintain necessary local dependencies, our method also preserves KV pairs within a sliding window (\Cref{fig:method}). Conversely, QA tasks exhibit sparse attention as they focus only on query-specific information. Consequently, gates trained solely on QA signals generalize poorly, failing to retain the sufficient information required for broader downstream tasks.

In \Crefsub{fig:opt-target}{b}, we study the choice of gate inputs by comparing hidden states, key states, and pre-RoPE key states \citep{rope}. Hidden states outperform the alternatives, indicating that they contain richer information for predicting the future utility of KV features. We observe that incorporating position-encoded features degrades performance. This motivates decoupling positional information from the gating mechanism and instead leveraging recency through a simple local-window strategy (\Cref{fig:method}).

\subsection{Architecture}\label{subsec:architecture}

\paragraph{Consideration.} The primary objective in designing the gate architecture is to minimize computational and memory overhead while accurately modeling attention behavior to predict the future utilization of each KV feature.

\paragraph{Low-Rank Sink Attention.}
We introduce a lightweight, attention-inspired gating mechanism, illustrated in \Cref{fig:gate}. Given a hidden state $\mathbf{h}\in\mathbb{R}^D$, we apply linear projections followed by weighted normalization \citep{qwen3} to obtain $\mathbf{k}\in\mathbb{R}^{H\times D'}$ and $\mathbf{q}\in\mathbb{R}^{G\times H\times D'}$, where $H$ is the number of original KV heads and $G$ denotes the original grouped-query size \citep{gqa}. Both representations are low-dimensional relative to the original KV features.

\begin{figure}[t]
    \centering
    \resizebox{1.02\linewidth}{!}{%
\begin{tikzpicture}

    \begin{groupplot}[
        group style={group name=groupa, columns=2, horizontal sep=1.2cm, vertical sep=0.0cm},
        width=4.0cm,
        height=3.5cm,
        every axis plot/.append style={thick},
        grid=major,
        xmajorgrids=true,
        ymajorgrids=true,
        major grid style={dotted, black},
        xlabel={KV cache ratio},
        ylabel={Accuracy (\%)},
        xlabel shift=-0.15cm,
        ylabel shift=-0.15cm,
        xlabel near ticks,
        ylabel near ticks,
        label style={font=\footnotesize},
        tick label style={font=\scriptsize},
        tick pos=left,
        xmax=1.0,
        xmin=0.2,
        xtick={0.2, 0.5, 0.75, 1.0},
        extra x ticks={0.3, 0.4},
        extra x tick labels={,},
        extra x tick style={grid=none, tick style={thin}},
    ]
        \nextgroupplot[
            y tick label style={/pgf/number format/.cd, fixed, fixed zerofill, precision=0},
            ytick distance=10,
            title={Retrieval (SCBench.KV)},
            title style={font=\small, yshift=-0.3em},
            legend columns=4,
            legend style={at={(1.12, 1.4)}, anchor=south, font=\footnotesize,
            /pgfplots/legend image code/.code={
                \draw[mark repeat=2,mark phase=2]
                plot coordinates {(0cm,0cm) (0.2cm,0cm) (0.4cm,0cm)};
            }},
        ]
        \addplot[gate] table [y=attn-retv, col sep=comma]{data/opt-architecture.csv};\addlegendentry{Ours}
        \addplot[duo] table [y=nosink-retv, col sep=comma]{data/opt-architecture.csv};\addlegendentry{Ours w/o $\mathbf{k}_{\text{sink}}$, $b_j$}
        \addplot[snap] table [y=mlp-retv, col sep=comma]{data/opt-architecture.csv};\addlegendentry{MLP}
        \addplot[expect] table [y=linear-retv, col sep=comma]{data/opt-architecture.csv};\addlegendentry{Linear}

        \nextgroupplot[
            y tick label style={/pgf/number format/.cd, fixed, fixed zerofill, precision=0},
            ymin=90,
            ytick distance=1.0,
            title={Contextual QA (SQuAD)},
            title style={font=\small, yshift=-0.3em},
        ]
        \addplot[gate] table [y=attn-qa, col sep=comma]{data/opt-architecture.csv};
        \addplot[duo] table [y=nosink-qa, col sep=comma]{data/opt-architecture.csv};
        \addplot[snap] table [y=mlp-qa, col sep=comma]{data/opt-architecture.csv};
        \addplot[expect] table [y=linear-qa, col sep=comma]{data/opt-architecture.csv};
    \end{groupplot}
    \node[below, font=\normalsize] at ($(groupa c1r1.south)!0.5!(groupa c2r1.south)+(0,-0.7cm)$) {(a) Effect of gate architectures};

    \begin{scope}[xshift=7.7cm]
    \begin{groupplot}[
        group style={group name=groupb, columns=2, horizontal sep=1.2cm, vertical sep=0.0cm},
        width=4cm,
        height=3.5cm,
        every axis plot/.append style={thick},
        grid=major,
        xmajorgrids=true,
        ymajorgrids=true,
        major grid style={dotted, black},
        xlabel={Peak KV Memory (GB)},
        xlabel shift=-0.15cm,
        ylabel shift=-0.15cm,
        xlabel near ticks,
        ylabel near ticks,
        label style={font=\footnotesize},
        tick label style={font=\scriptsize},
        tick pos=left,
        xmin=0.0,
        xmax=20.0,
    ]
        \nextgroupplot[
            ylabel={Time (s)},
            ymin=0.0,
            ytick distance=40,
            title={Prefill},
            title style={font=\small, yshift=-0.3em},
            legend columns=3,
            legend style={at={(1.12, 1.4)}, anchor=south, font=\footnotesize,
                /pgfplots/legend image code/.code={
                    \draw[mark repeat=2,mark phase=2]
                    plot coordinates {(0cm,0cm) (0.2cm,0cm) (0.4cm,0cm)};
                }},
        ]
        \addplot[gate] table [x=prefill-memory-gate, y=prefill-time-gate, col sep=comma]{data/profile-qwen2.5-7b.csv};\addlegendentry{Fast KVzip (ours)}
        \addplot[snap] table [x=prefill-memory-kvzip, y=prefill-time-kvzip, col sep=comma]{data/profile-qwen2.5-7b.csv};\addlegendentry{KVzip}
        \addplot[expect] table [x=prefill-memory-naive, y=prefill-time-naive, col sep=comma]{data/profile-qwen2.5-7b.csv};\addlegendentry{No Compression}

        \nextgroupplot[
            ylabel={Time per Token (ms)},
            ymin=0.0,
            ytick distance=10,
            title={Decoding},
            title style={font=\small, yshift=-0.3em},
        ]
        \addplot[gate] table [x=decode-memory-gate, y=time-token-gate, col sep=comma]{data/profile-qwen2.5-7b.csv};
        \addplot[expect] table [x=decode-memory-naive, y=time-token-naive, col sep=comma]{data/profile-qwen2.5-7b.csv};
    \end{groupplot}
    \node[below, font=\normalsize] at ($(groupb c1r1.south)!0.5!(groupb c2r1.south)+(0,-0.7cm)$) {(b) Inference efficiency};
    \end{scope}

\end{tikzpicture}
    }
    \caption{(a) \textbf{Architecture analysis}. We compare our sink-attention, a variant without learnable denominator terms, a two-layer MLP, and a linear model, using Qwen2.5-7B-1M. For a fair comparison, we equalize the number of parameters across all architectures. (b) \textbf{Prefill and decoding efficiency} of the Qwen2.5-7B-1M model using a 30\% KV budget ratio with PyTorch and FlashAttention-2 \citep{flashattn} on a single H100 GPU. Points in the plot correspond to context lengths of 160K, 240K, and 320K. In \Cref{appendix:end_latency}, we provide an end-to-end latency analysis.}
    \label{fig:opt-arch-efficiency}
\end{figure}

\begin{wrapfigure}[10]{r}{0.5\textwidth}
    \vspace{-2em}
    \centering


\resizebox{1.0\linewidth}{!}{
\begin{tikzpicture}

\definecolor{myblue}{RGB}{120, 120, 240}
\definecolor{mygreen}{RGB}{120, 180, 60}

\definecolor{myred}{RGB}{230, 140, 100}
\definecolor{myorange}{RGB}{240, 220, 120}

\tikzstyle{kvcache} = [draw=black, line width=0.4pt, rounded corners=.06cm, minimum width=0.8cm, minimum height=0.2cm]
\tikzstyle{arrow} = [-stealth, line width=0.2 mm]
\tikzstyle{patterns} = [pattern=north east lines, pattern color=black, draw=none, minimum width=0.2cm, minimum height=0.25cm, inner sep=0pt, anchor=east]
\tikzstyle{label} = [font=\scriptsize]
\tikzstyle{txt} = [font=\footnotesize]

\newcommand{\drawstack}[2]{
    \node[kvcache, fill=myorange, anchor=center, opacity=0.85] (#1_back) at ($(#2) + (0.15, 0.12)$) {};
    \node[kvcache, fill=myred, anchor=center, opacity=0.85] (#1_front) at (#2) {};
}

\newcommand{\drawtriplestack}[2]{
    \node[kvcache, fill=myorange, anchor=center, opacity=0.85] (#1_back2) at ($(#2) + (0.45, 0.36)$) {};
    \node[kvcache, fill=myorange, anchor=center, opacity=0.85] (#1_back1) at ($(#2) + (0.3, 0.24)$) {};
    \node[kvcache, fill=myred, anchor=center, opacity=0.85] (#1_front) at ($(#2) + (0.15, 0.12)$) {};
    \node[kvcache, fill=myred, anchor=center, opacity=0.85] (#1_front) at (#2) {};
}

\node[kvcache, fill=myblue!50, minimum width=2.0cm] (hidden) at (0,0) {};
\node[txt, left=0.cm of hidden] {Hidden state};

\coordinate (key_pos) at (0, 1.3);
\drawstack{key}{key_pos}
\node[txt, right=0.cm of key_front, yshift=-0.22cm] {$\mathbf{k}$};

\coordinate (query_pos) at ([xshift=1.5cm]key_pos);
\drawtriplestack{query}{query_pos}
\node[txt, right=0.cm of query_front, yshift=-0.22cm] {$\mathbf{q}$};

\draw[arrow] (hidden.north) -- (key_front.south);
\draw[arrow] (0, 0.86) -| (query_front.south);
\node[label, align=left, anchor=west] at (0.05, 0.56) {Projection (low rank)\\+ Normalization};


\coordinate (init1_pos) at ([xshift=-3.8cm]key_pos);
\coordinate (inits_pos) at ([xshift=-1.9cm]key_pos);

\node[fill=gray!8, rounded corners=.15cm, minimum width=3.8cm, minimum height=0.3cm, inner sep=0.5cm, left=0.5cm of key_pos] (shared_bg) {};

\drawstack{init1}{init1_pos}
\node[txt, right=0.cm of init1_front, yshift=-0.27cm] {$\mathbf{k}_{\text{sink}}^1$};

\node at ($ (init1_pos)!0.5!(inits_pos) $) {...};

\drawstack{inits}{inits_pos}
\node[txt, right=0.cm of inits_front, yshift=-0.27cm] {$\mathbf{k}_{\text{sink}}^S$};

\node[label,below=0.cm of shared_bg.south] {Parameters per layer};


\coordinate (top_line_y) at ($(key_back.north) + (0, 0.5)$);

\draw[] (init1_back.north |- top_line_y) -- (query_back1.north |- top_line_y);
\draw[] (query_back1.north |- top_line_y) -- ($(query_back1.north)+(0.,0.1)$);

\draw[arrow] (init1_back.north |- top_line_y) -- ($(init1_back.north)+(0.,0.1)$);
\draw[arrow] (inits_back.north |- top_line_y) -- ($(inits_back.north)+(0.,0.1)$);
\draw[arrow] (key_back.north |- top_line_y) -- ($(key_back.north)+(0.,0.1)$);


\node[above=0.8cm of key_pos] (arrow) {\Large $\Uparrow$}; 
\node[label, right=-0.18cm of arrow, yshift=-0.05cm] (att_label) {Attention weight};

\coordinate (scores_group_pos) at ([yshift=1.7cm]key_pos);
\begin{scope}[shift={(scores_group_pos)}]
    \node[kvcache, fill=myorange, minimum width=0.25cm, minimum height=0.25cm, opacity=0.85] at (0.15,0.12) {};
    \node[kvcache, fill=myorange, minimum width=0.25cm, minimum height=0.25cm, opacity=0.85] (s2) at (0,0) {};
    \node[kvcache, fill=myred, minimum width=0.25cm, minimum height=0.25cm, opacity=0.85] at (-0.15,-0.12) {};
    \node[kvcache, fill=myred, minimum width=0.25cm, minimum height=0.25cm, opacity=0.85] (s1) at (-0.3,-0.24) {};
\end{scope}

\coordinate (final_score_pos) at ([xshift=1.6cm]scores_group_pos);
\begin{scope}[shift={(final_score_pos)}]
    \node[kvcache, fill=myorange, minimum width=0.25cm, minimum height=0.25cm, opacity=0.85] (fs) at (0,0) {};
    \node[kvcache, fill=myred, minimum width=0.25cm, minimum height=0.25cm, opacity=0.85] at (-0.15,-0.12) {};
\end{scope}

\node[txt, right=0.cm of fs, yshift=-0.15cm] {Scores};

\draw[arrow] ($(s2.east) + (0.18,-0.1)$) -- ($(fs.west) - (0.18,0.1)$) node[label, midway, above] {Average};

\end{tikzpicture}
}

    \caption{\textbf{Gate architecture.} We propose a sink-attention architecture using a fixed set of keys per layer. Given a hidden state $\mathbf{h}\in\mathbb{R}^D$, we compute the importance scores of KV as $\mathbf{s}\in[0,1]^H$.}
    \label{fig:gate}
\end{wrapfigure}

The gate produces an importance score $\mathbf{s}\in\mathbb{R}^H$ via a sink-attention mechanism as
\vspace{-0.2em}
{\small
\begin{align*}
    \mathbf{s} = \frac{1}{G} \sum_{j=1}^G \frac{\text{exp}({\mathbf{q}_j}^\intercal\mathbf{k})}{\text{exp}({\mathbf{q}_j}^\intercal\mathbf{k})+ \underbrace{\sum_{r=1}^{S} \text{exp}({\mathbf{q}_j}^\intercal{\mathbf{k}_{\text{sink}}^{r}})}_{\text{Learnable sinks}}+b_j}.
\end{align*}
}
Here, $\mathbf{k}_{\text{sink}}^{r} \in \mathbb{R}^{H \times D'}$ are layer-specific learnable parameters that model sink keys, which play a critical role in standard attention mechanisms \citep{streaming}. We empirically observe that these features enable the gate to effectively estimate the importance of KV features for future attention. A scalar $b_j \ge 0$ serves as a bias controlling head-level importance. These parameters are trained jointly with the projection and normalization weights.

Unlike DeepSeek-V3.2, which employs a low-rank KV cache within its Lightning Indexer to enable sparse attention \citep{deepseekv3.2}, our method uses a fixed set of low-rank key features per layer and introduces no additional KV caching. This design simplifies implementation and enables straightforward integration into existing architectures.

Note that we use an identical set of hyperparameters across all models, setting $S=16$ and $D'=16$, which is $1/8$ of the Qwen3 head dimension \citep{qwen3}, which suggests that our method requires neither exhaustive hyperparameter search nor manual fine-tuning. We provide a sensitivity analysis of hyperparameters in \Cref{appendix:sensitivity_gate}.

\paragraph{Alternatives.} 
We study alternative gate architectures, including a linear model, a two-layer MLP with SwiGLU activation \citep{swiglu}, and a variant of our architecture without a learnable denominator, \textit{i.e.,} setting $\sum_{r=1}^{S} \exp({\mathbf{q}_j}^\intercal{\mathbf{k}_{\text{sink}}^{r}}) + b_j = 1$. Except for the linear model, we match the parameter counts across all architectures.
\Crefsub{fig:opt-arch-efficiency}{a} shows that while alternative architectures are competitive on contextual QA tasks, they exhibit degraded performance on retrieval tasks. These results highlight the effectiveness of our sink-attention gate architecture.

\subsection{Efficiency}
\paragraph{Gate Training.} 
\Cref{tab:efficiency} shows the training time and memory requirements of our gate training process, demonstrating its efficiency across diverse model scales.

\begin{wrapfigure}[10]{r}{0.5\textwidth}
    \vspace{-1.2em}
    \centering
    \captionof{table}{\textbf{Training efficiency.} Gate training time using a H100 GPU and gate sizes across diverse model scales. The 30B-A3B model has a smaller hidden state size than the other models, resulting in smaller gates.}
    \centering
    \small
    \begin{tabular}{lrr}
    \toprule
    Model    & \hspace{-1em}Training Time (h)  &  Storage (GB)  \\
    \midrule
    Qwen3-8B    & 0.59 & 0.18 \\
    Qwen3-30B-A3B  & 0.70  & 0.11 \\
    Qwen3-14B  &  0.83 & 0.30 \\
    \bottomrule
    \end{tabular}
    \label{tab:efficiency}   
\end{wrapfigure}

\paragraph{Inference.} \Crefsub{fig:opt-arch-efficiency}{b} shows the memory and latency efficiency improvements of Fast KVzip during both prefill and decoding. The results demonstrate that Fast KVzip reduces latency and memory usage compared to the base model, while significantly improving prefilling efficiency compared to KVzip by removing the runtime KV cache compression overhead. For all experiments, we apply a non-uniform KV cache eviction structure following \citet{kvzip}.

\section{Experiment}\label{sec:exp}

We evaluate Fast KVzip under both prefill-intensive and decoding-intensive scenarios. Note, we train gates once per model using FineWeb-Edu \citep{fineweb} and evaluate on all downstream tasks (see \Cref{subsec:method_optimization}), demonstrating that the learned gates are task-generalizable. All gate training and model evaluations are conducted on a single NVIDIA H100 80GB GPU using PyTorch and FlashAttention-2 \citep{flashattn}.

\subsection{Setup}

\paragraph{Datasets.}
For prefill-intensive scenarios, we evaluate all methods on SCBench \citep{scbench}, a multi-query benchmark comprising nine tasks, including synthetic key retrieval tasks from RULER \citep{ruler}, long-context question answering from LongBench \citep{longbench}, and code comprehension tasks. We further include OpenAI MRCR \citep{mrcr}, a multi-round conversational retrieval benchmark, and SQuAD \citep{squad}, a multi-query QA dataset.  The maximum context length reaches up to 170K tokens (Qwen3 tokenizer).

For decoding-intensive tasks, we evaluate on mathematical reasoning benchmarks, AIME24 \citep{aime} and MATH \citep{math}. Following the R-KV baseline \citep{rkv}, we set the maximum decoding length to 32K tokens for AIME24 and 16K tokens for MATH.

\paragraph{Models.}
We conduct experiments on state-of-the-art open-source LLMs, including Qwen2.5-7B/14B-1M \citep{qwen}, Qwen3-8B/14B, Qwen3-8B-FP8 \citep{qwen3}, and Gemma3-12B \citep{gemma3}. We evaluate each model using its native precision. Notably, Gemma3-12B employs a hybrid sliding-window attention mechanism, while Qwen3-8B-FP8 adopts dynamic FP8 quantization. For the hybrid model, we only compress the global attention KV cache, which dominates memory usage in long-context evaluation 

\paragraph{Baselines.}
We compare Fast KVzip against several state-of-the-art KV cache compression methods, including KVzip \citep{kvzip}, SnapKV \citep{snapkv}, Expected Attention \citep{kvpress} with Ada-KV \citep{adakv}, DuoAttention \citep{duo} with KVzip score \citep{kvzip}, R-KV \citep{rkv}, and TrimKV \citep{trimkv}. For prefill-intensive tasks, we evaluate methods under the query-agnostic setting, following the setting of KVzip \citep{kvzip}. 

For decoding-intensive tasks, we adopt the R-KV setting \citep{rkv}, which continuously compresses the KV cache during decoding, covering both the context and generated tokens, to match the fixed-size cache budget. Additionally, we evaluate the early stopping of thinking strategy introduced in Qwen3 \citep{qwen3}, where the model terminates the thinking process upon reaching a predefined budget and is forced to generate the final answer. Since some baselines are specifically designed for either prefill-intensive or decoding-intensive scenarios, we report comparisons only within their respective applicable settings.

\begin{figure*}[t]
    \centering
    \begin{tikzpicture}

\begin{groupplot}[group style={columns=4, rows=3, horizontal sep=1.08cm, vertical sep=1.2cm},
    width=3.9cm,
    height=3.4cm,
    every axis plot/.append style={thick},
    xlabel shift=-0.12cm,         
    ylabel shift=-0.16cm,
    xlabel near ticks,
    ylabel near ticks,
    label style={font=\scriptsize},
    xlabel={KV cache ratio},
    grid=major,
    xmajorgrids=true,
    ymajorgrids=true,
    major grid style={dotted, black},
    tick label style={font=\scriptsize},
    tick pos=left,
    y tick label style={/pgf/number format/.cd, fixed, fixed zerofill, precision=0},
    ytick distance=20,
    xmax=1.0,
    xmin=0.2,
    xtick={0.2, 0.5, 0.75, 1.0},
    extra x ticks={0.3, 0.4},   
    extra x tick labels={,},
    extra x tick style={
        grid=none,
        tick style={thin},
    },    
    title style={
      at={(axis description cs:0.5,0.88)}, 
      font={\footnotesize},
      text height=1.5ex,
      text depth=0.25ex,
    },    
]


\nextgroupplot[title=OpenAI MRCR, ylabel={Accuracy (\%)}, ytick distance=5, ymax=30]
\addplot[kvzip] table[x=ratio, y=kvzip-mrcr, col sep=comma]{data/qwen2.5-7b.csv};
\addplot[duo] table[x=ratio, y=duo-mrcr, col sep=comma]{data/qwen2.5-7b.csv};
\addplot[snap] table[x=ratio, y=snap-mrcr, col sep=comma]{data/qwen2.5-7b.csv};
\addplot[expect] table[x=ratio, y=expect-mrcr, col sep=comma]{data/qwen2.5-7b.csv};
\addplot[gate] table[x=ratio, y=gate-mrcr, col sep=comma]{data/qwen2.5-7b.csv};

\nextgroupplot[title=Retr.KV, ylabel={Accuracy (\%)}, ymax=80]
\addplot[kvzip] table[x=ratio, y=kvzip-kv, col sep=comma]{data/qwen2.5-7b.csv};
\addplot[duo] table[x=ratio, y=duo-kv, col sep=comma]{data/qwen2.5-7b.csv};
\addplot[snap] table[x=ratio, y=snap-kv, col sep=comma]{data/qwen2.5-7b.csv};
\addplot[expect] table[x=ratio, y=expect-kv, col sep=comma]{data/qwen2.5-7b.csv};
\addplot[gate] table[x=ratio, y=gate-kv, col sep=comma]{data/qwen2.5-7b.csv};

\nextgroupplot[title=Retr.Prefix-Suffix, ylabel={Accuracy (\%)}, ytick distance=15, ymax=60,
legend columns=5, legend style={at={(2.4,1.3)}, anchor=south east, font=\footnotesize},
]
\addplot[gate] table[x=ratio, y=gate-prefix, col sep=comma]{data/qwen2.5-7b.csv};\addlegendentry{Fast KVzip (ours)}
\addplot[duo] table[x=ratio, y=duo-prefix, col sep=comma]{data/qwen2.5-7b.csv};\addlegendentry{DuoAttention}
\addplot[snap] table[x=ratio, y=snap-prefix, col sep=comma]{data/qwen2.5-7b.csv};\addlegendentry{SnapKV}
\addplot[expect] table[x=ratio, y=expect-prefix, col sep=comma]{data/qwen2.5-7b.csv};\addlegendentry{Expected Attention (+ AdaKV)}
\addplot[kvzip] table[x=ratio, y=kvzip-prefix, col sep=comma]{data/qwen2.5-7b.csv};\addlegendentry{KVzip}

\nextgroupplot[title=Code.RepoQA, ylabel={Pass@1 (\%)}, ymin=0]
\addplot[kvzip] table[x=ratio, y=kvzip-repoqa, col sep=comma]{data/qwen2.5-7b.csv};
\addplot[duo] table[x=ratio, y=duo-repoqa, col sep=comma]{data/qwen2.5-7b.csv};
\addplot[snap] table[x=ratio, y=snap-repoqa, col sep=comma]{data/qwen2.5-7b.csv};
\addplot[expect] table[x=ratio, y=expect-repoqa, col sep=comma]{data/qwen2.5-7b.csv};
\addplot[gate] table[x=ratio, y=gate-repoqa, col sep=comma]{data/qwen2.5-7b.csv};


\nextgroupplot[title=SQuAD, ylabel={Accuracy (\%)}, ymax=100, ymin=20]
\addplot[kvzip] table[x=ratio, y=kvzip-squad, col sep=comma]{data/qwen2.5-7b.csv};
\addplot[duo] table[x=ratio, y=duo-squad, col sep=comma]{data/qwen2.5-7b.csv};
\addplot[snap] table[x=ratio, y=snap-squad, col sep=comma]{data/qwen2.5-7b.csv};
\addplot[expect] table[x=ratio, y=expect-squad, col sep=comma]{data/qwen2.5-7b.csv};
\addplot[gate] table[x=ratio, y=gate-squad, col sep=comma]{data/qwen2.5-7b.csv};

\nextgroupplot[title=GSM8K, ylabel={Accuracy (\%)}, ymax=80, ymin=0]
\addplot[kvzip] table[x=ratio, y=kvzip-gsm, col sep=comma]{data/qwen2.5-7b.csv};
\addplot[duo] table[x=ratio, y=duo-gsm, col sep=comma]{data/qwen2.5-7b.csv};
\addplot[snap] table[x=ratio, y=snap-gsm, col sep=comma]{data/qwen2.5-7b.csv};
\addplot[expect] table[x=ratio, y=expect-gsm, col sep=comma]{data/qwen2.5-7b.csv};
\addplot[gate] table[x=ratio, y=gate-gsm, col sep=comma]{data/qwen2.5-7b.csv};

\nextgroupplot[title=En.QA, ylabel={Accuracy (\%)}, ytick distance=8, ymin=16, ymax=48]
\addplot[kvzip] table[x=ratio, y=kvzip-qa, col sep=comma]{data/qwen2.5-7b.csv};
\addplot[duo] table[x=ratio, y=duo-qa, col sep=comma]{data/qwen2.5-7b.csv};
\addplot[snap] table[x=ratio, y=snap-qa, col sep=comma]{data/qwen2.5-7b.csv};
\addplot[expect] table[x=ratio, y=expect-qa, col sep=comma]{data/qwen2.5-7b.csv};
\addplot[gate] table[x=ratio, y=gate-qa, col sep=comma]{data/qwen2.5-7b.csv};

\nextgroupplot[title=En.MultiChoice, ylabel={Accuracy (\%)}, , ytick distance=8, ymax=88]
\addplot[kvzip] table[x=ratio, y=kvzip-choice, col sep=comma]{data/qwen2.5-7b.csv};
\addplot[duo] table[x=ratio, y=duo-choice, col sep=comma]{data/qwen2.5-7b.csv};
\addplot[snap] table[x=ratio, y=snap-choice, col sep=comma]{data/qwen2.5-7b.csv};
\addplot[expect] table[x=ratio, y=expect-choice, col sep=comma]{data/qwen2.5-7b.csv};
\addplot[gate] table[x=ratio, y=gate-choice, col sep=comma]{data/qwen2.5-7b.csv};


\nextgroupplot[title=En.Summary, ylabel={ROUGE (\%)}, ytick distance=4, ymin=24]
\addplot[kvzip] table[x=ratio, y=kvzip-summary, col sep=comma]{data/qwen2.5-7b.csv};
\addplot[duo] table[x=ratio, y=duo-summary, col sep=comma]{data/qwen2.5-7b.csv};
\addplot[snap] table[x=ratio, y=snap-summary, col sep=comma]{data/qwen2.5-7b.csv};
\addplot[expect] table[x=ratio, y=expect-summary, col sep=comma]{data/qwen2.5-7b.csv};
\addplot[gate] table[x=ratio, y=gate-summary, col sep=comma]{data/qwen2.5-7b.csv};

\nextgroupplot[title=Retr.MultiHop, ylabel={Accuracy (\%)}, ytick distance=10, ymax=50]
\addplot[kvzip] table[x=ratio, y=kvzip-vt, col sep=comma]{data/qwen2.5-7b.csv};
\addplot[duo] table[x=ratio, y=duo-vt, col sep=comma]{data/qwen2.5-7b.csv};
\addplot[snap] table[x=ratio, y=snap-vt, col sep=comma]{data/qwen2.5-7b.csv};
\addplot[expect] table[x=ratio, y=expect-vt, col sep=comma]{data/qwen2.5-7b.csv};
\addplot[gate] table[x=ratio, y=gate-vt, col sep=comma]{data/qwen2.5-7b.csv};

\nextgroupplot[title=Math.Find, ylabel={Accuracy (\%)}, ytick distance=5]
\addplot[kvzip] table[x=ratio, y=kvzip-mf, col sep=comma]{data/qwen2.5-7b.csv};
\addplot[duo] table[x=ratio, y=duo-mf, col sep=comma]{data/qwen2.5-7b.csv};
\addplot[snap] table[x=ratio, y=snap-mf, col sep=comma]{data/qwen2.5-7b.csv};
\addplot[expect] table[x=ratio, y=expect-mf, col sep=comma]{data/qwen2.5-7b.csv};
\addplot[gate] table[x=ratio, y=gate-mf, col sep=comma]{data/qwen2.5-7b.csv};

\nextgroupplot[title=ICL.ManyShot, ylabel={Accuracy (\%)}, ytick distance=4, ymax=40, ymin=28]
\addplot[kvzip] table[x=ratio, y=kvzip-manyshot, col sep=comma]{data/qwen2.5-7b.csv};
\addplot[duo] table[x=ratio, y=duo-manyshot, col sep=comma]{data/qwen2.5-7b.csv};
\addplot[snap] table[x=ratio, y=snap-manyshot, col sep=comma]{data/qwen2.5-7b.csv};
\addplot[expect] table[x=ratio, y=expect-manyshot, col sep=comma]{data/qwen2.5-7b.csv};
\addplot[gate] table[x=ratio, y=gate-manyshot, col sep=comma]{data/qwen2.5-7b.csv};

\end{groupplot}

\node[rotate=90, align=center, anchor=center, font=\bfseries\footnotesize] at ($(group c1r1.west)+(-1.15cm,0)$) {Retrieval};
\node[rotate=90, align=center, anchor=center, font=\bfseries\footnotesize] at ($(group c1r2.west)+(-1.15cm,0)$) {Contextual QA};
\node[rotate=90, align=center, anchor=center, font=\bfseries\footnotesize] at ($(group c1r3.west)+(-1.15cm,0)$) {Redundancy};

\end{tikzpicture}
    \vspace{-1.5em}
    \caption{\textbf{Prefill-intensive benchmark results} with Qwen2.5-7B-1M under varying KV cache budget ratios from 0.2 to 1.0. Tasks are organized into three categories: (1) retrieval-intensive, (2) contextual understanding, and (3) high context redundancy.}
    \label{graph:prefill_benchmark}
    \vspace{1.5em}
    \centering
    \begin{tikzpicture}

\begin{groupplot}[group style={columns=4, horizontal sep=1.2cm, vertical sep=1.2cm},
    width=3.9cm,
    height=3.4cm,
    every axis plot/.append style={thick},
    xlabel shift=-0.08cm,         
    ylabel shift=-0.15cm,
    xlabel near ticks,
    ylabel near ticks,
    label style={font=\scriptsize},
    xlabel={KV cache ratio},
    ylabel={Rel. performance},
    grid=major,
    xmajorgrids=true,
    ymajorgrids=true,
    major grid style={dotted, black},
    tick label style={font=\scriptsize},
    tick pos=left,
    y tick label style={/pgf/number format/.cd, fixed, fixed zerofill, precision=1},
    xmax=1.0,
    xmin=0.2,
    xtick={0.2, 0.5, 0.75, 1.0},
    extra x ticks={0.3, 0.4},   
    extra x tick labels={,},
    extra x tick style={
        grid=none,
        tick style={thin},
    },    
    extra y ticks={0.3, 0.5, 0.7, 0.9},   
    extra y tick labels={,,,},
    extra y tick style={
        grid=none,
        tick style={thin},
        major tick length=2.4pt,
    },    
    title style={
      at={(axis description cs:0.5,0.88)}, 
      anchor=south,
      font={\footnotesize},
      text height=1.5ex,
      text depth=0.25ex,
    },    
]


\nextgroupplot[title=Qwen3-8B,  
legend columns=5, legend style={at={(-0.1,1.35)}, anchor=south west, font=\footnotesize},
]
\addplot[gate] table[x=ratio, y=gate-qwen3-8b, col sep=comma]{data/architecture.csv};\addlegendentry{Fast KVzip (ours)}
\addplot[duo] table[x=ratio, y=duo-qwen3-8b, col sep=comma]{data/architecture.csv};\addlegendentry{DuoAttention}
\addplot[snap] table[x=ratio, y=snap-qwen3-8b, col sep=comma]{data/architecture.csv};\addlegendentry{SnapKV}
\addplot[expect] table[x=ratio, y=expect-qwen3-8b, col sep=comma]{data/architecture.csv};\addlegendentry{Expected Attention (+ AdaKV)}
\addplot[kvzip] table[x=ratio, y=kvzip-qwen3-8b, col sep=comma]{data/architecture.csv};\addlegendentry{KVzip}

\nextgroupplot[title=Qwen2.5-14B-1M]
\addplot[kvzip] table[x=ratio, y=kvzip-qwen2.5-14b, col sep=comma]{data/architecture.csv};
\addplot[duo] table[x=ratio, y=duo-qwen2.5-14b, col sep=comma]{data/architecture.csv};
\addplot[snap] table[x=ratio, y=snap-qwen2.5-14b, col sep=comma]{data/architecture.csv};
\addplot[expect] table[x=ratio, y=expect-qwen2.5-14b, col sep=comma]{data/architecture.csv};
\addplot[gate] table[x=ratio, y=gate-qwen2.5-14b, col sep=comma]{data/architecture.csv};

\nextgroupplot[title=Gemma3-12B, xlabel={KV cache ratio (global)}]
\addplot[kvzip] table[x=ratio, y=kvzip-gemma3-12b, col sep=comma]{data/architecture.csv};
\addplot[duo] table[x=ratio, y=duo-gemma3-12b, col sep=comma]{data/architecture.csv};
\addplot[snap] table[x=ratio, y=snap-gemma3-12b, col sep=comma]{data/architecture.csv};
\addplot[expect] table[x=ratio, y=expect-gemma3-12b, col sep=comma]{data/architecture.csv};
\addplot[gate] table[x=ratio, y=gate-gemma3-12b, col sep=comma]{data/architecture.csv};

\nextgroupplot[title=Qwen3-8B-FP8]
\addplot[kvzip] table[x=ratio, y=kvzip-qwen3-8b-fp8, col sep=comma]{data/architecture.csv};
\addplot[duo] table[x=ratio, y=duo-qwen3-8b-fp8, col sep=comma]{data/architecture.csv};
\addplot[snap] table[x=ratio, y=snap-qwen3-8b-fp8, col sep=comma]{data/architecture.csv};
\addplot[expect] table[x=ratio, y=expect-qwen3-8b-fp8, col sep=comma]{data/architecture.csv};
\addplot[gate] table[x=ratio, y=gate-qwen3-8b-fp8, col sep=comma]{data/architecture.csv};

\end{groupplot}
\end{tikzpicture}
    \vspace{-1.5em}
    \caption{\textbf{Performance across different models}, averaged over 12 benchmark datasets. For each dataset, results are normalized with respect to full-cache performance prior to averaging.}
    \label{graph:architecture}
\end{figure*}

\subsection{Prefill-Intensive Tasks}

\Cref{graph:prefill_benchmark} shows the prefill KV cache compression performance across 12 datasets. The results demonstrate that Fast KVzip outperforms all baselines across all tasks while matching the performance of KVzip, which requires twice the prefilling computational cost for compression. The figure shows that Fast KVzip maintains full-cache performance at a 30$\minus$40\% KV cache budget ratio. On some tasks, Fast KVzip exhibits performance improvements, which we attribute to the denoising effect of attention induced by KV cache compression \citep{differential}.

\Cref{graph:architecture} presents the averaged relative performance across the 12 datasets for various LLMs, demonstrating that Fast KVzip maintains effectiveness across models with a different pre-/post-training (Qwen3-8B), a larger model scale (Qwen2.5-14B-1M), a different attention mechanism (Gemma3-12B), and a quantization (Qwen3-8B-FP8).

\subsection{Decoding-Intensive Tasks}

\begin{figure*}[t]
  \begin{center}
    \begin{tikzpicture}

\tikzstyle{kvzip} = [purple!70!black, mark=x, mark size=1.4pt, densely dashed]

\begin{groupplot}[group style={columns=4, horizontal sep=1.2cm, vertical sep=1.2cm},
    width=3.9cm,
    height=3.4cm,
    every axis plot/.append style={thick},
    xlabel shift=-0.08cm,         
    ylabel shift=-0.15cm,
    xlabel near ticks,
    ylabel near ticks,
    label style={font=\scriptsize},
    xlabel={KV cache budget},
    ylabel={Accuracy (\%)},
    grid=major,
    xmajorgrids=true,
    ymajorgrids=true,
    major grid style={dotted, black},
    tick label style={font=\scriptsize},
    tick pos=left,
    x tick label style={/pgf/number format/.cd, fixed, fixed zerofill, precision=0},
    y tick label style={/pgf/number format/.cd, fixed, fixed zerofill, precision=0},
    extra x tick style={
        grid=none,
        tick style={thin},
    },    
    extra y ticks={30, 50, 70},   
    extra y tick labels={,,,},
    extra y tick style={
        grid=none,
        tick style={thin},
        major tick length=2.4pt,
    },    
    title style={
      at={(axis description cs:0.5,0.88)}, 
      anchor=south,
      font={\footnotesize}
    },    
]


\nextgroupplot[title=AIME24 (Qwen3-8B),  
    legend columns=5, legend style={at={(0.55,1.35)}, anchor=south west, font=\footnotesize},
    xtick={2048, 4096, 6144},
    extra x ticks={3072},   
    extra x tick labels={},
    ymax=80,
]
\addplot[gate] table[x=budget, y=gate-8b, col sep=comma]{data/aime.csv};\addlegendentry{Fast KVzip (ours)}
\addplot[duo] table[x=budget, y=rkv-8b, col sep=comma]{data/aime.csv};\addlegendentry{R-KV}
\addplot[snap] table[x=budget, y=snap-8b, col sep=comma]{data/aime.csv};\addlegendentry{SnapKV}
\addplot[expect] table[x=budget, y=early-8b, col sep=comma]{data/aime.csv};\addlegendentry{Early Stopping}
\addplot[kvzip] table[x=budget, y=full-8b, col sep=comma]{data/aime.csv};\addlegendentry{Full KV}

\nextgroupplot[title=AIME24 (Qwen3-14B),
    xtick={2048, 4096, 6144},
    extra x ticks={3072},   
    extra x tick labels={},
]
\addplot[gate] table[x=budget, y=gate-14b, col sep=comma]{data/aime.csv};
\addplot[duo] table[x=budget, y=rkv-14b, col sep=comma]{data/aime.csv};
\addplot[snap] table[x=budget, y=snap-14b, col sep=comma]{data/aime.csv};
\addplot[expect] table[x=budget, y=early-14b, col sep=comma]{data/aime.csv};
\addplot[kvzip] table[x=budget, y=full-14b, col sep=comma]{data/aime.csv};

\nextgroupplot[title=MATH (Qwen3-8B),
    xtick={2048, 3072, 4096},
]
\addplot[gate] table[x=budget, y=gate-8b, col sep=comma]{data/math.csv};
\addplot[duo] table[x=budget, y=rkv-8b, col sep=comma]{data/math.csv};
\addplot[snap] table[x=budget, y=snap-8b, col sep=comma]{data/math.csv};
\addplot[expect] table[x=budget, y=early-8b, col sep=comma]{data/math.csv};
\addplot[kvzip] table[x=budget, y=full-8b, col sep=comma]{data/math.csv};

\nextgroupplot[title=MATH (Qwen3-14B), 
    xtick={2048, 3072, 4096},
]
\addplot[gate] table[x=budget, y=gate-14b, col sep=comma]{data/math.csv};
\addplot[duo] table[x=budget, y=rkv-14b, col sep=comma]{data/math.csv};
\addplot[snap] table[x=budget, y=snap-14b, col sep=comma]{data/math.csv};
\addplot[expect] table[x=budget, y=early-14b, col sep=comma]{data/math.csv};
\addplot[kvzip] table[x=budget, y=full-14b, col sep=comma]{data/math.csv};

\end{groupplot}
\end{tikzpicture}
    \vspace{-1.5em}
    \caption{\textbf{Decoding-intensive tasks}. For AIME24, we averaged scores over 16 random seeds.}
    \label{graph:decoding_benchmark}
  \end{center}
\end{figure*}

\Cref{graph:decoding_benchmark} shows the KV cache compression performance on mathematical reasoning tasks, including AIME24 and MATH. The results demonstrate that Fast KVzip achieves superior performance compared to decoding-scenario compression baselines on Qwen3-8B and Qwen3-14B models. Fast KVzip achieves near-lossless performance with a KV budget size of 4K, whereas the early-stopping-of-thinking method drastically reduces reasoning performance. These results indicate that the model requires a sufficient length of thinking to deduce correct answers, and that Fast KVzip indeed maintains the necessary KV features for general inference, including reasoning procedures.

\section{Qualitative Analysis}\label{sec:analysis}

\begin{wrapfigure}[12]{r}{0.5\textwidth}
    \vspace{-2em}
    \centering
    \begin{tikzpicture}
    \begin{axis}[
        width=6cm, 
        height=1.3cm,
        scale only axis, enlargelimits=false,
        axis on top,
        xlabel={Layer ($0\rightarrow27$)},
        ylabel={Head},        
        label style={font=\footnotesize},
        xlabel shift=-0.1cm,
        ylabel shift=-0.1cm,
        axis line style={draw=none},
        tick style={draw=none},
        xtick=\empty, 
        ytick=\empty,        
        colorbar,
        colorbar style={
            height=1.1cm,
            width=0.2cm,
            at={(parent axis.east)},
            anchor=west,
            xshift=-0.1cm,
            font=\footnotesize,
            yshift=0cm,
            ylabel={},
            ytick={0, 1.0},     
        },        colormap/viridis,
        point meta min=0.0,     
        point meta max=1.0,     
    ]
        \addplot graphics [
            xmin=0, xmax=1, 
            ymin=0, ymax=1
        ] {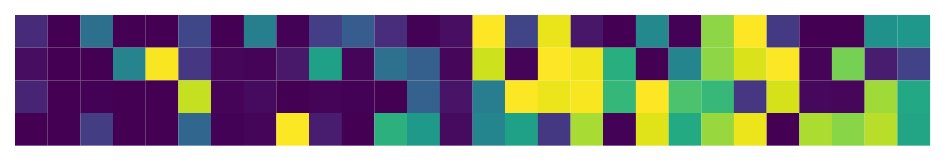};        
    \end{axis}
\end{tikzpicture}
    \vspace{-1em}
    \caption{\textbf{Retention Rates per Attention Head.} We apply a 36\% KV cache budget to Qwen2.5-7B-1M, which has 28 layers with 4 KV heads per layer. The left side corresponds to the early layers. Heads in the middle layers tend to retain the majority of token information, whereas gates in the early layers learn to retain only a small subset of KV features.}
    \label{fig:heatmap-pred}
\end{wrapfigure}

We analyze the gating behavior of Qwen2.5-7B-1M using English book and code data from SCBench. \Cref{fig:heatmap-pred} illustrates the distribution of KV cache retention rates across attention heads under a total of 36\% KV cache budget. We observe that half of the heads (primarily in early layers) exhibit highly sparse KV caches, while the remaining heads show uniformly distributed retention rates. This suggests that the gates prioritize high-level features from middle-to-late layers over low-level features from early layers.

We categorize attention heads by their retention rates ($r$) into three types: sparse ($r<0.05$), medium ($0.05\le r<0.9$), and dense ($0.9\le r$). \Cref{tab:head_group} details the tokens most frequently evicted and retained by each head type. Notably, sparse heads predominantly retain tokens associated with punctuation or line-break tokens. In contrast, other heads tend to retain tokens that are frequent in the source data; these frequent tokens appear as both the most retained and most evicted in medium heads. 
This complementary behavior across head types ensures that low-frequency but semantically important entities are preserved by the dense and medium heads, while sparse heads handle structural tokens, capturing high-level summarized information \citep{punctuation} which indicates an attention pattern distinct from naive local-window mechanisms \citep{streaming}. 

\begin{table}[t]
    \centering
    \caption{\textbf{Gate behavior analysis.} The most frequently retained and evicted tokens for each head type. The {\textbackslash}s represents a space.}
    \vspace{0.2em}
    \small
    \begin{tabular}{l|r|r}
    \toprule
    Head Type & Most Retained (Top $1 \rightarrow 8$) & Most Evicted (Top $1 \rightarrow 8$)\\
    \midrule
    Sparse & Structural markers (. .{\textbackslash}n{\textbackslash}n .{\textbackslash}n , :{\textbackslash}n {\textbackslash}n {\textbackslash}s)& High-frequency tokens (, the . and to of a I)\\
    Medium & High-frequency tokens (, . and the of to I in) & High-frequency tokens (the , and to of a . I) \\
    Dense & High-frequency tokens (, the . and to of a I)& Whitespace tokens (the {\textbackslash}s , a {\textbackslash}t {\textbackslash}s{\textbackslash}s ( 1)\\
    \bottomrule
    \end{tabular}  
    \label{tab:head_group}
\end{table}
\section{Conclusion}
We propose Fast KVzip, a gating-based KV cache compression method that matches state-of-the-art performance while eliminating runtime compression overhead. Through principled analysis, we design an effective gating optimization and a lightweight sink-attention architecture. Extensive evaluations on both prefill- and decoding-intensive tasks show that Fast KVzip generalizes well across diverse downstream applications, including retrieval, code comprehension, and reasoning, reducing KV cache size to 30\% with negligible performance loss.


\section*{Acknowledgement}
We thank the NAVER AI Lab team for valuable discussions. 
Jang-Hyun led the project and developed the main idea. 
Dongyoon and Sangdoo provided senior supervision and contributed to the methodology and experiments. 
Jang-Hyun and Sangdoo are corresponding authors.


\bibliography{main}
\bibliographystyle{abbrvnat}

\newpage
\appendix
\newpage
\appendix
\onecolumn


\section{Experimental Details}\label{appendix:details}
In this section, we detail the training procedure in \Cref{fig:optimization}, describe the training hyperparameters used in \Cref{subsec:method_optimization}, and the gate architecture detailed in \Cref{subsec:architecture}.

\subsection{Gate Training Target Score}\label{appendix:target}
In \Cref{fig:optimization}, we construct the input as ``<text> Repeat the previous context <text>" for each text sample from our training data, following KVzip \citep{kvzip}. Through the LLM forward pass, the repeated text attends to the keys of the original text. For simplicity, let us assume the number of attention heads as 1. From the attention matrix $A^l$ at layer $l$, we compute importance score of each key states of the original text by the maximum attention it receives:
$$s_j^l = \max_{i \in \text{repeated}} \; A^l[i, j],$$
where $i$ ranges over the repeated text positions and $j$ indexes keys in the original text. Intuitively, a key that no repeated query attends to is safely evictable, while a key that at least one query strongly attends to should be retained.

We then train a lightweight gate $g^l$ at each layer to predict these scores directly given the corresponding input hidden states $h_j^l$. Given target scores $s_j^l$, we optimize the gate using binary cross-entropy loss:
$$\mathcal{L}^l = -\sum_j \left[ s_j^l \log g^l(h_j^l) + (1 - s_j^l) \log (1 - g^l(h_j^l)) \right],$$
while freezing all LLM weights. The optimization is embarrassingly parallel across both samples and layers, enabling efficient GPU execution.

\subsection{Dataset}\label{appendix:dataset}
As described in \Cref{subsec:method_optimization}, we use the FineWeb-Edu dataset \citep{fineweb} to train the gates. We randomly subsample sequences with context lengths ranging from 10K to 30K tokens, for a total of 500K tokens. By concatenating these samples, we construct an additional long-context dataset, yielding sequences of length 100K tokens and totaling another 500K tokens. In total, we use 1M tokens for gate training.

\subsection{Training Hyperparameters}\label{appendix:hyperparameter}
For efficiency, we first precompute tuples of hidden states and their corresponding KV importance scores using the dataset described above. We then optimize the gates in parallel using stochastic gradient descent with a binary cross-entropy loss and a learning rate of 0.2 across all experiments. Training is performed for 5K update steps with a batch size of 1K. Given the total of 1M training tokens, this corresponds to 5 epochs over the dataset.

\Cref{fig:training_loss} illustrates the training loss trajectory and demonstrates stable convergence. We observe that earlier layers tend to exhibit lower loss, whereas the middle and later layers show higher loss values. We conjecture that higher layers perform more complex attention mechanisms, which are inherently more difficult to predict from a single hidden state. 

\Cref{fig:diff-kvzip} presents a token-level comparison of predictions, illustrating that our trained gating mechanism captures token-level dynamics and avoids over-smoothing.

\begin{figure}[ht]
    \centering
    \resizebox{\linewidth}{!}{%
    \begin{tikzpicture}

    \begin{groupplot}[group style={columns=2, horizontal sep=2cm, vertical sep=0.0cm},
            width=8.4cm,
            height=6cm,
            every axis plot/.append style={thick},
            grid=major,
            xmajorgrids=true,
            ymajorgrids=true,
            major grid style={dotted, black},
            xlabel={Update step},
            ylabel={Loss},
            xlabel shift=-0.15cm,
            ylabel shift=-0.15cm,
            xlabel near ticks,
            ylabel near ticks,
            label style={font=\footnotesize},
            tick label style={font=\scriptsize},
            tick pos=left,
            xmax=5000.0,
            xmin=0.0,
            ymax=2.5,
            ymin=0.0,
            ytick distance=0.5,
            legend style={legend columns=2, font=\scriptsize},
            legend cell align={left},
    ]
        \nextgroupplot[
            no marks, 
            thick,
            y tick label style={/pgf/number format/.cd, fixed, fixed zerofill, precision=1},
            title={(a) Training loss},
            title style={font=\small, yshift=-0.4em},
        ]

        \addplot+ [red] table [x=steps, y=layer0, col sep=comma] {data/loss_train.csv};\addlegendentryexpanded{Layer 0-3}
        \addplot+ [or] table [x=steps, y=layer1, col sep=comma] {data/loss_train.csv};\addlegendentryexpanded{Layer 4-7}
        \addplot+ [yl] table [x=steps, y=layer2, col sep=comma] {data/loss_train.csv};\addlegendentryexpanded{Layer 8-11}
        \addplot+ [gr] table [x=steps, y=layer3, col sep=comma] {data/loss_train.csv};\addlegendentryexpanded{Layer 12-15}
        \addplot+ [sky] table [x=steps, y=layer4, col sep=comma] {data/loss_train.csv};\addlegendentryexpanded{Layer 16-19}
        \addplot+ [bl] table [x=steps, y=layer5, col sep=comma] {data/loss_train.csv};\addlegendentryexpanded{Layer 20-23}
        \addplot+ [pp] table [x=steps, y=layer6, col sep=comma] {data/loss_train.csv};\addlegendentryexpanded{Layer 24-27}

        \nextgroupplot[
            no marks, 
            thick,
            y tick label style={/pgf/number format/.cd, fixed, fixed zerofill, precision=1},
            title={(b) Validation loss},
            title style={font=\small, yshift=-0.4em},
        ]

        \addplot+ [red] table [x=steps, y=layer0, col sep=comma] {data/loss_val.csv};\addlegendentryexpanded{Layer 0-3}
        \addplot+ [or] table [x=steps, y=layer1, col sep=comma] {data/loss_val.csv};\addlegendentryexpanded{Layer 4-7}
        \addplot+ [yl] table [x=steps, y=layer2, col sep=comma] {data/loss_val.csv};\addlegendentryexpanded{Layer 8-11}
        \addplot+ [gr] table [x=steps, y=layer3, col sep=comma] {data/loss_val.csv};\addlegendentryexpanded{Layer 12-15}
        \addplot+ [sky] table [x=steps, y=layer4, col sep=comma] {data/loss_val.csv};\addlegendentryexpanded{Layer 16-19}
        \addplot+ [bl] table [x=steps, y=layer5, col sep=comma] {data/loss_val.csv};\addlegendentryexpanded{Layer 20-23}
        \addplot+ [pp] table [x=steps, y=layer6, col sep=comma] {data/loss_val.csv};\addlegendentryexpanded{Layer 24-27}

    \end{groupplot}
\end{tikzpicture}
    }
    \caption{Gate training and validation loss trajectory per attention layer for Qwen2.5-7B-1M. We plot loss values averaged every four layers in the model.}
    \label{fig:training_loss}
\end{figure}

\subsection{Gate Configuration}\label{appendix:gate}
As described in \Cref{subsec:architecture}, each gate uses 16 $\mathbf{k}_{\text{sink}}^s$ vectors. The low-rank projection dimension $D'$ is set to 16 for all experiments.

\subsection{Inference Details}\label{appendix:window}
During prefill, we employ chunked prefill with a chunk size of 16K tokens. As illustrated in \Cref{fig:method}, KV cache eviction is performed after each prefill chunk, while retaining KV features from the most recent 4K-token window to exploit recency bias in attention. For contexts shorter than 16K tokens, we retain the last 2\% of tokens to accommodate arbitrary input lengths. 

During decoding, KV eviction begins once the cache reaches the predefined budget. We use a hidden cache of size 128 and compute gating decisions in parallel. After updating importance scores, we evict an equivalent number of KV entries to maintain the cache budget. A local attention window of 128 tokens is preserved at each eviction step during decoding.

\section{Additional Experiments}\label{appendix:sensitivity}

\subsection{Training Data Source}
\label{appendix:data_source}

To analyze how training data affects gate behavior, we train a gate on OpenR1-Math data \citep{openr1} with the identical training configuration, and evaluated on AIME24 \citep{aime}. As shown in \Cref{table:training_data}, the math-domain gate performs comparably to our general-corpus gate. This result is intuitive: since our reconstruction-based objective trains the gate to preserve the context itself rather than to learn task-specific features, the choice of training domain has minimal impact, rather the broad coverage of FineWeb-Edu yields better generalization. This explains why our approach generalizes well across diverse tasks, while next-token prediction-based methods like TrimKV \citep{trimkv} exhibit strong performance specifically on the domain of their training data.

\begin{table}[h]
\centering
\caption{Effect of gate training data on AIME24 (Qwen3-8B).}
\vspace{0.2em}
\small
\begin{tabular}{lrrrr}
\toprule
KV Cache Budget & Fast KVzip (OpenR1 Math) & Fast KVzip (FineWeb-Edu) & SnapKV & R-KV \\
\midrule
4096 & 70.6\% & 72.7\% & 63.7\% & 60.0\% \\
2048 & 53.3\% & 63.1\% & 42.3\% & 32.3\% \\
512 & 16.7\% & 13.3\% & 0.8\% & 0.0\% \\
\bottomrule
\end{tabular}
\label{table:training_data}
\end{table}

\subsection{Training Data Size}
\Cref{graph:hyper-data} shows the performance of gates trained with different sizes of FineWeb-Edu data, as described in \Cref{appendix:dataset}. The figure demonstrates that increasing the size of the training data leads to performance improvements, particularly on SCBench.KV, the synthetic key-retrieval task. To further evaluate the effect of long-context data, we construct the 500K-short dataset by removing the long-context samples (each with a context length of 100K) from the 1M-token dataset described in \Cref{appendix:dataset}. The results demonstrate the effectiveness of incorporating long-context data for gate training.

\begin{figure}[ht]
    \centering
    \resizebox{\linewidth}{!}{%
    \begin{tikzpicture}

    \begin{groupplot}[group style={columns=2, horizontal sep=2cm, vertical sep=0.0cm},
            width=8cm,
            height=5.4cm,
            every axis plot/.append style={thick},
            grid=major,
            xmajorgrids=true,
            ymajorgrids=true,
            major grid style={dotted, black},
            xlabel={KV cache ratio},
            ylabel={Accuracy (\%)},
            xlabel shift=-0.15cm,
            ylabel shift=-0.15cm,
            xlabel near ticks,
            ylabel near ticks,
            label style={font=\footnotesize},
            tick label style={font=\scriptsize},
            tick pos=left,
            xmax=1.0,
            xmin=0.2,
            xtick={0.2, 0.3, 0.4, 0.5, 0.75, 1.0},
            legend image post style={scale=1.0},
            legend style={legend columns=1, font=\footnotesize, at={(0.98,0.04)}, anchor=south east},
            legend cell align={left},
    ]
        \nextgroupplot[
            y tick label style={/pgf/number format/.cd, fixed, fixed zerofill, precision=0},
            ytick distance=10,
            title={(a) Retrieval (SCBench.KV)},
            title style={font=\small, yshift=-0.4em},
        ]

        \addplot[gate] table [y=1M-kv, col sep=comma]{data/hyper-data.csv};\addlegendentry{1M}
        \addplot[duo] table [y=500K-kv, col sep=comma]{data/hyper-data.csv};\addlegendentry{500K}
        \addplot[snap] table [y=short-kv, col sep=comma]{data/hyper-data.csv};\addlegendentry{500K-short}
        \addplot[expect] table [y=200K-kv, col sep=comma]{data/hyper-data.csv};\addlegendentry{200K}
        \addplot[pp] table [y=50K-kv, col sep=comma]{data/hyper-data.csv};\addlegendentry{50K}

        \nextgroupplot[
            y tick label style={/pgf/number format/.cd, fixed, fixed zerofill, precision=0},
            ytick distance=1,
            title={(b) Contextual QA (SQuAD)},
            title style={font=\small, yshift=-0.4em},
        ]

        \addplot[gate] table [y=1M, col sep=comma]{data/hyper-data.csv};\addlegendentry{1M}
        \addplot[duo] table [y=500K, col sep=comma]{data/hyper-data.csv};\addlegendentry{500K}
        \addplot[snap] table [y=short, col sep=comma]{data/hyper-data.csv};\addlegendentry{500K-short}
        \addplot[expect] table [y=200K, col sep=comma]{data/hyper-data.csv};\addlegendentry{200K}
        \addplot[pp] table [y=50K, col sep=comma]{data/hyper-data.csv};\addlegendentry{50K}

    \end{groupplot}
\end{tikzpicture}
    }
    \caption{\textbf{Effect of Training Data Size.} We evaluate gates trained with different amounts of training data sampled from FineWeb-Edu using Qwen2.5-7B-1M. For the other settings, we uniformly reduce the context lengths of the data.}
    \label{graph:hyper-data}
\end{figure}

\subsection{End-to-End Latency}\label{appendix:end_latency}
We provide end-to-end results on SCBench-KV, the longest-context benchmark in our evaluation ($\sim$170K tokens). As shown in \Cref{tab:end_to_end_latency}, KVzip incurs nearly 3$\times$ end-to-end latency compared to vanilla due to its reconstruction overhead. FastKVzip eliminates this overhead entirely, achieving up to 5$\times$ speedup over KVzip at 30\% KV ratio while maintaining comparable accuracy. This speedup stems from two sources: (1) gating removes the per-query reconstruction cost, and (2) chunked prefill compression (\Cref{fig:method}) further accelerates the prefill stage itself.

\begin{table}[h]
\centering
\caption{End-to-end latency and accuracy on SCBench-KV with Qwen2.5-7B-1M. Acceleration ratios in parentheses are relative to vanilla (ratio 1.0). FastKVzip eliminates reconstruction overhead entirely, achieving up to 5$\times$ speedup over KVzip while maintaining comparable accuracy.}
\vspace{0.3em}
\small
\begin{tabular}{l|rr|rr}
\toprule
KV Ratio & FastKVzip Time (s) & KVzip Time (s) & FastKVzip Acc & KVzip Acc \\
\midrule
1.00 & 22.5 (1.0$\times$) & 69.1 & 68.2 & 68.2 \\
0.50 & 16.8 (1.3$\times$) & 68.7 & 71.6 & 68.6 \\
0.40 & 15.0 (1.5$\times$) & 68.4 & 66.4 & 66.0 \\
0.30 & 13.1 (1.7$\times$) & 68.2 & 65.4 & 65.0 \\
\bottomrule
\end{tabular}
\label{tab:end_to_end_latency}
\end{table}

\subsection{Gate Configuration}\label{appendix:sensitivity_gate}
\Cref{graph:hyper-dim} analyzes the effects of the number of sink keys $S$ and the projection dimension $D'$ in our sink-attention architecture (\Cref{subsec:architecture}). The figure shows that performance is robust to the choice of projection dimension, while highlighting the importance of having a sufficient number of sink keys for effective KV cache compression.

\begin{figure}[ht]
    \centering
    \resizebox{\linewidth}{!}{%
    \begin{tikzpicture}

    \begin{groupplot}[group style={columns=2, horizontal sep=2cm, vertical sep=0.0cm},
            width=8cm,
            height=5.4cm,
            every axis plot/.append style={thick},
            grid=major,
            xmajorgrids=true,
            ymajorgrids=true,
            major grid style={dotted, black},
            xlabel={KV cache ratio},
            ylabel={Accuracy (\%)},
            xlabel shift=-0.15cm,
            ylabel shift=-0.15cm,
            xlabel near ticks,
            ylabel near ticks,
            label style={font=\footnotesize},
            tick label style={font=\scriptsize},
            tick pos=left,
            xmax=1.0,
            xmin=0.2,
            xtick={0.2, 0.3, 0.4, 0.5, 0.75, 1.0},
            legend image post style={scale=1.0},
            legend style={legend columns=1, font=\footnotesize, at={(0.98,0.04)}, anchor=south east},
            legend cell align={left},
    ]
        \nextgroupplot[
            y tick label style={/pgf/number format/.cd, fixed, fixed zerofill, precision=0},
            ymin=40.0,
            title={(a) Projection Dimension Size},
            title style={font=\small, yshift=-0.4em},
        ]

        \addplot[gate] table [y=mid-kv, col sep=comma]{data/hyper-dim.csv};\addlegendentry{$D'=16$}
        \addplot[snap] table [y=high-kv, col sep=comma]{data/hyper-dim.csv};\addlegendentry{$D'=32$}
        \addplot[expect] table [y=low-kv, col sep=comma]{data/hyper-dim.csv};\addlegendentry{$D'=8$}

        \nextgroupplot[
            y tick label style={/pgf/number format/.cd, fixed, fixed zerofill, precision=0},
            title={(b) The Number of Sink Keys},
            title style={font=\small, yshift=-0.4em},
        ]

        \addplot[gate] table [y=16-kv, col sep=comma]{data/hyper-sink.csv};\addlegendentry{$S=16$}
        \addplot[duo] table [y=64-kv, col sep=comma]{data/hyper-sink.csv};\addlegendentry{$S=64$}
        \addplot[snap] table [y=4-kv, col sep=comma]{data/hyper-sink.csv};\addlegendentry{$S=4$}
        \addplot[expect] table [y=1-kv, col sep=comma]{data/hyper-sink.csv};\addlegendentry{$S=1$}
        \addplot[pp] table [y=0-kv, col sep=comma]{data/hyper-sink.csv};\addlegendentry{$S=0$}

    \end{groupplot}
\end{tikzpicture}
    }
    \caption{\textbf{Effect of Gate Configuration.} We evaluate gates with different numbers of sink keys $S$ and projection dimensions $D'$ using Qwen2.5-7B-1M on SCBench.KV.}
    \label{graph:hyper-dim}
\end{figure}

\subsection{Inference Configuration}\label{appendix:exp_local_window}
\Cref{graph:hyper-window} shows the effect of the local window size in our inference algorithm described in \Cref{fig:method}. The figure demonstrates that retaining a local window improves performance, while performance is comparable across local window sizes ranging from 0K (no local window) to 8K tokens.

\begin{figure}[ht]
    \centering
    \resizebox{\linewidth}{!}{%
    \begin{tikzpicture}

    \begin{groupplot}[group style={columns=2, horizontal sep=2cm, vertical sep=0.0cm},
            width=8cm,
            height=5.4cm,
            every axis plot/.append style={thick},
            grid=major,
            xmajorgrids=true,
            ymajorgrids=true,
            major grid style={dotted, black},
            xlabel={KV cache ratio},
            ylabel={Accuracy (\%)},
            xlabel shift=-0.15cm,
            ylabel shift=-0.15cm,
            xlabel near ticks,
            ylabel near ticks,
            label style={font=\footnotesize},
            tick label style={font=\scriptsize},
            tick pos=left,
            xmax=1.0,
            xmin=0.2,
            xtick={0.2, 0.3, 0.4, 0.5, 0.75, 1.0},
            legend image post style={scale=1.0},
            legend style={legend columns=1, font=\footnotesize, at={(0.98,0.04)}, anchor=south east},
            legend cell align={left},
    ]
        \nextgroupplot[
            y tick label style={/pgf/number format/.cd, fixed, fixed zerofill, precision=0},
            ymin=40.0,
            title={(a) Retrieval (SCBench.KV)},
            title style={font=\small, yshift=-0.4em},
        ]

        \addplot[gate] table [y=4k-kv, col sep=comma]{data/hyper-window.csv};\addlegendentry{4K}
        \addplot[duo] table [y=8k-kv, col sep=comma]{data/hyper-window.csv};\addlegendentry{8K}
        \addplot[snap] table [y=2k-kv, col sep=comma]{data/hyper-window.csv};\addlegendentry{2K}
        \addplot[expect] table [y=1k-kv, col sep=comma]{data/hyper-window.csv};\addlegendentry{1K}
        \addplot[pp] table [y=0w-kv, col sep=comma]{data/hyper-window.csv};\addlegendentry{0K}

        \nextgroupplot[
            y tick label style={/pgf/number format/.cd, fixed, fixed zerofill, precision=0},
            ymin=30,
            title={(b) Contextual QA (SCBench.En.QA)},
            title style={font=\small, yshift=-0.4em},
        ]

        \addplot[gate] table [y=4k, col sep=comma]{data/hyper-window.csv};\addlegendentry{4K}
        \addplot[duo] table [y=8k, col sep=comma]{data/hyper-window.csv};\addlegendentry{8K}
        \addplot[snap] table [y=2k, col sep=comma]{data/hyper-window.csv};\addlegendentry{2K}
        \addplot[expect] table [y=1k, col sep=comma]{data/hyper-window.csv};\addlegendentry{1K}
        \addplot[pp] table [y=0w, col sep=comma]{data/hyper-window.csv};\addlegendentry{0K}

    \end{groupplot}
\end{tikzpicture}
    }
    \caption{\textbf{Effect of Local Window Size.} We evaluate gates with a range of local window sizes, as described in \Cref{fig:method}, using Qwen2.5-7B-1M. Note that our algorithm performs eviction  that all configurations have an identical total KV cache size after eviction. In our main experiment, set the local window hyperparameter as 4K tokens for prefill and 128 tokens for decoding.}
    \label{graph:hyper-window}
\end{figure}

\section{Extended Comparison with Prior Gating Approaches}\label{appendix:experiment}

In \Cref{tab:gating_comparison}, we provide a comparison with prior gating-based approaches for KV cache compression, including Locret \citep{locret}, DMS \citep{dms}, and TrimKV \citep{trimkv}. 

In \Cref{graph:trimkv}, we further evaluate our method on Qwen3-4B-Instruct-2507, for which an official TrimKV model for general language tasks is available. For other released models, TrimKV trains specialized gating models for mathematical reasoning using math-specific datasets, which prevents a fair comparison with our method, trained on general data in a task-agnostic manner. As shown in \Cref{graph:trimkv}, our Fast KVzip outperforms TrimKV in general task settings, particularly on retrieval tasks. These results demonstrate the effectiveness of our gating optimization and design, emphasizing their importance.

\begin{table}[ht]
    \caption{\textbf{Comparison of gating approaches} for KV cache compression. While DMS and TrimKV trains separate gates for reasoning and instruction-following tasks using distinct datasets, we employ a single dataset for gate training across all models and evaluations.}
    \label{tab:gating_comparison}
    \centering
    \resizebox{\linewidth}{!}{
    \begin{tabular}{lllll}
    \toprule
    Method & Training Data & Task & Target & Gate Architecture \\
    \midrule
    Locret \citep{locret} & Instruction-based QA & Question-answering & Attention scores & MLP \\
    DMS \citep{dms} & Mathematics (or Programming) & Next-token logit distillation & End-to-end loss & Linear \\
    TrimKV \citep{trimkv} & Mathematics (or Book) & Next-token prediction & End-to-end loss & MLP \\
    \midrule
    Fast KVzip (Ours) & General text data & Reconstruction & Attention scores & Sink attention \\
    \bottomrule
    \end{tabular}
    }
\end{table}

\begin{figure}[ht]
    \centering
    \resizebox{\linewidth}{!}{%
    \begin{tikzpicture}

    \begin{groupplot}[group style={columns=2, horizontal sep=2cm, vertical sep=0.0cm},
            width=8cm,
            height=5.2cm,
            every axis plot/.append style={thick},
            grid=major,
            xmajorgrids=true,
            ymajorgrids=true,
            major grid style={dotted, black},
            xlabel={KV cache ratio},
            ylabel={Rel. performance},
            xlabel shift=-0.15cm,
            ylabel shift=-0.15cm,
            xlabel near ticks,
            ylabel near ticks,
            label style={font=\footnotesize},
            tick label style={font=\scriptsize},
            tick pos=left,
            xmax=1.0,
            xmin=0.2,
            xtick={0.2, 0.5, 0.75, 1.0},
            extra x ticks={0.3, 0.4},   
            extra x tick labels={0.3, 0.4},
            extra x tick style={
                grid=none,
                tick style={thin},
            },    
            legend image post style={scale=1.0},
            legend style={legend columns=1, font=\footnotesize, at={(0.98,0.08)}, anchor=south east},
            legend cell align={left},
    ]
        \nextgroupplot[
            y tick label style={/pgf/number format/.cd, fixed, fixed zerofill, precision=1},
            ymin=0.15,
            title={(a) Retrieval},
            title style={font=\small, yshift=-0.4em},
        ]

        \addplot[gate] table [y=gate-retv, col sep=comma]{data/trimkv.csv};\addlegendentry{Fast KVzip}
        \addplot[expect] table [y=trimkv-retv, col sep=comma]{data/trimkv.csv};\addlegendentry{TrimKV}

        \nextgroupplot[
            y tick label style={/pgf/number format/.cd, fixed, fixed zerofill, precision=2},
            ymin=0.9,
            ytick distance=0.02,
            title={(b) Contextual QA},
            title style={font=\small, yshift=-0.4em},
        ]

        \addplot[gate] table [y=gate-qa, col sep=comma]{data/trimkv.csv};\addlegendentry{Fast KVzip}
        \addplot[expect] table [y=trimkv-qa, col sep=comma]{data/trimkv.csv};\addlegendentry{TrimKV}

    \end{groupplot}
\end{tikzpicture}
    }
    \caption{\textbf{Comparison to TrimKV} \citep{trimkv} with prefill-intensive tasks. We evaluate the official TrimKV model with Qwen3-4B-Instruct-2507 on retrieval and contextual QA tasks in \Cref{graph:prefill_benchmark}, and report the averaged scores. For each dataset, results are normalized with respect to full-cache performance prior to averaging. For Locret, the released models are outdated, and we replace performance comparisons with our analysis in \Cref{fig:opt-target}.}
    \label{graph:trimkv}
\end{figure}

\begin{figure}[!ht]
  \begin{center}
    \resizebox{\linewidth}{!}{
    \begin{tikzpicture}
    \begin{groupplot}[
        group style={
            group size=4 by 7,      
            horizontal sep=1cm,   
            vertical sep=0.7cm,     
            x descriptions at=edge bottom,  
        },
        width=4.6cm, 
        height=3.5cm,        
        every axis plot/.append style={thick},
        grid=major,
        no markers,
        xmajorgrids=true,
        ymajorgrids=true,
        major grid style={dotted, black},
        title style={yshift=-0.7em},
        xlabel={KV sequence},
        ylabel={},        
        scaled y ticks=false, 
        y tick label style={/pgf/number format/fixed},
        xtick pos=bottom, 
        ytick pos=left,
        xtick={0, 50, 100},
        xticklabels={0, 40K, 80K},
        xmax=114,
        xmin=-6,
    ]
    \pgfplotsinvokeforeach{0,...,26}{
        \pgfmathsetmacro{\isLeftColumn}{int(mod(#1,4)==1 ? 1 : 0)}        
        
        \nextgroupplot[
            title={Layer #1}, 
            ylabel={\ifnum\isLeftColumn=1 Score\fi},
        ]
        \addplot[black, dashed, thick, domain=-4:110] {0.03};
        
        \addplot+[thin,blue] table [x=seq, y=target#1, col sep=comma] {data/pred.csv};
        
         \addplot+[thin,red] table [x=seq, y=pred#1, col sep=comma] {data/pred.csv};
    }

    \nextgroupplot[
        title={Layer 27}, 
        legend to name=CommonLegend,
        legend columns=3,            
    ]
    \addplot+[thin, red] table [x=seq, y=pred27, col sep=comma] {data/pred.csv};
    \addlegendentry{Fast KVzip}

    \addplot+[thin, blue] table [x=seq, y=target27, col sep=comma] {data/pred.csv};
    \addlegendentry{KVzip}

    \addplot[black, dashed, thick, domain=-4:110] {0.03};\addlegendentry{Threshold (0.03)}

    \end{groupplot}
    
    \node[anchor=south] at ($(group c2r1.north)!0.5!(group c3r1.north) + (0, 0.8cm)$) {
        \ref{CommonLegend}
    };

\end{tikzpicture}
    }
    \caption{\textbf{Prediction Comparison of Fast KVzip to KVzip.} We use Qwen2.5-7B-1M with the SCBench.En.QA dataset, which were not used during training. For each layer, we plot the averaged score over heads. The dashed line represents the threshold value that results in a 36\% KV cache budget ratio.}
    \label{fig:diff-kvzip} 
  \end{center}
\end{figure}

\section{Broader Impacts and Limitations}\label{appendix:limitation}
\paragraph{Broader impacts.}
This paper explores technical methods to enhance LLM efficiency and does not present direct social risks. Socially, our findings have a positive impact by decreasing the costs associated with deploying and maintaining large-scale models. By minimizing resource consumption, this work facilitates more sustainable and cost-effective AI operations for the broader community.

\paragraph{Limitations and Future Work.}
Prior gating methods for attention modules, such as MoD \citep{mixture-depth} and MoBA \citep{moba}, typically require from-scratch training to fit predefined structures. An important future direction is to study how applying our method during pretraining can induce hardware-efficient structured computation patterns. Moreover, exploring a unified framework that integrates these gating approaches is a promising avenue. In particular, the gating mechanism could be extended to support multi-choice decisions, including skipping computation, discarding KV features, or storing them for selective retrieval.


\newpage
\section*{NeurIPS Paper Checklist}

The checklist is designed to encourage best practices for responsible machine learning research, addressing issues of reproducibility, transparency, research ethics, and societal impact. Do not remove the checklist: {\bf The papers not including the checklist will be desk rejected.} The checklist should follow the references and follow the (optional) supplemental material.  The checklist does NOT count towards the page
limit. 

Please read the checklist guidelines carefully for information on how to answer these questions. For each question in the checklist:
\begin{itemize}
    \item You should answer \answerYes{}, \answerNo{}, or \answerNA{}.
    \item \answerNA{} means either that the question is Not Applicable for that particular paper or the relevant information is Not Available.
    \item Please provide a short (1--2 sentence) justification right after your answer (even for \answerNA). 
\end{itemize}

{\bf The checklist answers are an integral part of your paper submission.} They are visible to the reviewers, area chairs, senior area chairs, and ethics reviewers. You will also be asked to include it (after eventual revisions) with the final version of your paper, and its final version will be published with the paper.

The reviewers of your paper will be asked to use the checklist as one of the factors in their evaluation. While \answerYes{} is generally preferable to \answerNo{}, it is perfectly acceptable to answer \answerNo{} provided a proper justification is given (e.g., error bars are not reported because it would be too computationally expensive'' or ``we were unable to find the license for the dataset we used''). In general, answering \answerNo{} or \answerNA{} is not grounds for rejection. While the questions are phrased in a binary way, we acknowledge that the true answer is often more nuanced, so please just use your best judgment and write a justification to elaborate. All supporting evidence can appear either in the main paper or the supplemental material, provided in appendix. If you answer \answerYes{} to a question, in the justification please point to the section(s) where related material for the question can be found.

IMPORTANT, please:
\begin{itemize}
    \item {\bf Delete this instruction block, but keep the section heading ``NeurIPS Paper Checklist"},
    \item  {\bf Keep the checklist subsection headings, questions/answers and guidelines below.}
    \item {\bf Do not modify the questions and only use the provided macros for your answers}.
\end{itemize}


\begin{enumerate}

\item {\bf Claims}
    \item[] Question: Do the main claims made in the abstract and introduction accurately reflect the paper's contributions and scope?
    \item[] Answer: \answerYes{} 
    \item[] Justification: In the abstract and the introduction (especially, the third paragraph), we describe our scope as KV cache compression and state our technical contributions.

\item {\bf Limitations}
    \item[] Question: Does the paper discuss the limitations of the work performed by the authors?
    \item[] Answer: \answerYes{} 
    \item[] Justification: In Appendix D.

\item {\bf Theory assumptions and proofs}
    \item[] Question: For each theoretical result, does the paper provide the full set of assumptions and a complete (and correct) proof?
    \item[] Answer: \answerNA{} 
    \item[] Justification: We do not provide theoretical results.

    \item {\bf Experimental result reproducibility}
    \item[] Question: Does the paper fully disclose all the information needed to reproduce the main experimental results of the paper to the extent that it affects the main claims and/or conclusions of the paper (regardless of whether the code and data are provided or not)?
    \item[] Answer: \answerYes{} 
    \item[] Justification: Section 3 and Appendix A.

\item {\bf Open access to data and code}
    \item[] Question: Does the paper provide open access to the data and code, with sufficient instructions to faithfully reproduce the main experimental results, as described in supplemental material?
    \item[] Answer: \answerYes{} Data / \answerNo{} Code
    \item[] Justification: We use open-sourced training data (Appendix A) and evaluation data (Section 4). We plan to release our code in the camera-ready version.

\item {\bf Experimental setting/details}
    \item[] Question: Does the paper specify all the training and test details (e.g., data splits, hyperparameters, how they were chosen, type of optimizer) necessary to understand the results?
    \item[] Answer: \answerYes{} 
    \item[] Justification: Appendix A.

\item {\bf Experiment statistical significance}
    \item[] Question: Does the paper report error bars suitably and correctly defined or other appropriate information about the statistical significance of the experiments?
    \item[] Answer: \answerYes{} 
    \item[] Justification: We follow the convention. We run multiple seeds and report averaged score for math reasoning (AIME24, Section 4 Figure 11).

\item {\bf Experiments compute resources}
    \item[] Question: For each experiment, does the paper provide sufficient information on the computer resources (type of compute workers, memory, time of execution) needed to reproduce the experiments?
    \item[] Answer: \answerYes{} 
    \item[] Justification: Section 3.5 and Section 4.
    
\item {\bf Code of ethics}
    \item[] Question: Does the research conducted in the paper conform, in every respect, with the NeurIPS Code of Ethics \url{https://neurips.cc/public/EthicsGuidelines}?
    \item[] Answer: \answerYes{} 
    \item[] Justification: Yes, we follow the NeurIPS Code of Ethics.

\item {\bf Broader impacts}
    \item[] Question: Does the paper discuss both potential positive societal impacts and negative societal impacts of the work performed?
    \item[] Answer: \answerYes{} 
    \item[] Justification: Appendix D.
    
\item {\bf Safeguards}
    \item[] Question: Does the paper describe safeguards that have been put in place for responsible release of data or models that have a high risk for misuse (e.g., pre-trained language models, image generators, or scraped datasets)?
    \item[] Answer: \answerNA{} 
    \item[] Justification: N/A

\item {\bf Licenses for existing assets}
    \item[] Question: Are the creators or original owners of assets (e.g., code, data, models), used in the paper, properly credited and are the license and terms of use explicitly mentioned and properly respected?
    \item[] Answer: \answerNA{} 
    \item[] Justification: We use open-sourced models and datasets valid for the research purposes.

\item {\bf New assets}
    \item[] Question: Are new assets introduced in the paper well documented and is the documentation provided alongside the assets?
    \item[] Answer: \answerNA{} 
    \item[] Justification: N/A

\item {\bf Crowdsourcing and research with human subjects}
    \item[] Question: For crowdsourcing experiments and research with human subjects, does the paper include the full text of instructions given to participants and screenshots, if applicable, as well as details about compensation (if any)? 
    \item[] Answer: \answerNA{} 
    \item[] Justification: N/A

\item {\bf Institutional review board (IRB) approvals or equivalent for research with human subjects}
    \item[] Question: Does the paper describe potential risks incurred by study participants, whether such risks were disclosed to the subjects, and whether Institutional Review Board (IRB) approvals (or an equivalent approval/review based on the requirements of your country or institution) were obtained?
    \item[] Answer: \answerNA{} 
    \item[] Justification: N/A

\item {\bf Declaration of LLM usage}
    \item[] Question: Does the paper describe the usage of LLMs if it is an important, original, or non-standard component of the core methods in this research? Note that if the LLM is used only for writing, editing, or formatting purposes and does \emph{not} impact the core methodology, scientific rigor, or originality of the research, declaration is not required.
    \item[] Answer: \answerNA{} 
    \item[] Justification: We used the LLM for editing of our writing. 

\end{enumerate}

\end{document}